\documentclass[twoside,11pt]{article}

\usepackage[accepted]{melba}
\usepackage{color}
\usepackage{booktabs}
\usepackage{multirow}
\usepackage{colortbl}
\usepackage{appendix}
\usepackage{amsmath,amsfonts}

\usepackage{mwe} 

\usepackage{bm}
\usepackage{tabularx,booktabs,multirow,makecell,array}
\melbaheading{2022:027}{https://www.melba-journal.org/papers/2022:027.html}{2022}{1-33}{2021/12/8}{2022/11/15}{Zhao and Pang et al.}{}{}

\ShortHeadings{PU learning for cell detection in histopathology images with incomplete annotations}{Zhao and Pang et al.}
\firstpageno{1}

\title{Positive-unlabeled learning for binary and multi-class cell detection in histopathology images with incomplete annotations}

\author{\name Zipei Zhao$^{\mathrm{1}}$ (co-first author) \email 3120190728@bit.edu.cn \\  
	\addr School of Integrated Circuits and Electronics, Beijing Institute of Technology, Beijing, China.
	\AND
	\name Fengqian Pang$^{\mathrm{1}}$ (co-first author) \email fqpang@ncut.edu.cn \\
	\addr School of Information Science and Technology, North China University of Technology, Beijing, China.
	\AND
	\name Yaou Liu \email liuyaou@bjtth.org \\
	\addr Department of Radiology, Beijing Tiantan Hospital, Capital Medical University, Beijing, China.
	\AND
	\name Zhiwen Liu$^{\mathrm{*}}$ (co-corresponding author) \email zwliu@bit.edu.cn \\
	\addr School of Integrated Circuits and Electronics, Beijing Institute of Technology, Beijing, China.
	\AND
	\name Chuyang Ye$^{\mathrm{*}}$ (co-corresponding author) \email chuyang.ye@bit.edu.cn \\
	\addr School of Integrated Circuits and Electronics, Beijing Institute of Technology, Beijing, China.
}

\begin{document}

\maketitle

\begin{abstract}
\setlength{\marginparwidth}{0.2\textwidth}Cell detection in histopathology images is of great interest to clinical practice and research, and \textit{convolutional neural networks} (CNNs) have achieved remarkable cell detection results. Typically, to train CNN-based cell detection models, every positive instance in the training images needs to be annotated, and instances that are not labeled as positive are considered negative samples. 
However, manual cell annotation is complicated due to the large number and diversity of cells, and it can be difficult to ensure the annotation of every positive instance. 
In many cases, only incomplete annotations are available, where some of the positive instances are annotated and the others are not, and the classification loss term for negative samples in typical network training becomes incorrect.
In this work, to address this problem of incomplete annotations, we propose to reformulate the training of the detection network as a positive-unlabeled learning problem.
Since the instances in unannotated regions can be either positive or negative, they have unknown labels.
Using the samples with unknown labels and the positively labeled samples, we first derive an approximation of the classification loss term corresponding to negative samples for binary cell detection, and based on this approximation we further extend the proposed framework to multi-class cell detection.
For evaluation, experiments were performed on four publicly available datasets. The experimental results show that our method improves the performance of cell detection in histopathology images given incomplete annotations for network training.

\end{abstract}


\reversemarginpar
\section{Introduction}
With the continuous breakthrough of biological microscopic imaging technology, a large number of histopathology images have been produced to assist clinical practice and research. 
Quantitative, objective, and effective cell analysis based on histopathology images has become an important research direction~\citep{gurcan2009histopathological,veta2014breast}.
Pathologists usually use information such as the number, density, and distribution of cells in a given area in histopathology images to assess the degree of tissue damage and make a diagnosis~\citep{Fusi2013}. Such analysis relies on the detection of the cells of interest. However, manual cell detection performed by pathologists can be time-consuming and error-prone \citep{Wang2022,Laak2021}, especially in areas of high cell density, and automated cell detection methods are needed. 

In recent years, deep learning techniques have been successfully applied to various image processing tasks, and they have been increasingly used to analyze histopathology images as well~\citep{Lu2021AI,Noorbakhsh2020,SRINIDHI2021,HE2021,Laak2021,Marostica2021}. In particular, \textit{convolutional neural networks}~(CNNs) have been applied to perform automated cell detection in histopathology images.
For example, \cite{xu2015Stacked} propose stacked sparse autoencoder for efficient nuclei detection in high-resolution histopathology images of breast cancer; \cite{K2016Locality} propose a spatially constrained CNN to perform nuclei detection in routine colon cancer histopathology images. 
More advanced networks that are originally developed for generic object detection are later used or adapted for cell detection. In these methods, cells of interest are localized by bounding boxes.
For example, \cite{Cai2019} have modified Faster R-CNN~\citep{Ren2015Faster} for automatic mitosis detection in breast histopathology images; \cite{Sun2020} use the region proposal network~\citep{Ren2015Faster}, Faster R-CNN, and RetinaNet~\citep{Lin2017RetinaNet}, as well as their adapted versions that enable similarity learning, for signet ring cell detection in histopathology images.

To train advanced CNN-based cell detection models, usually every cell of interest in the training images should be annotated (e.g., with a bounding box and the identification of its type), and instances in unannotated regions\footnote{ Usually large whole-slide images are acquired for histopathology image analysis, and they are cropped into patches for cell annotation or detection. Here, the unannotated regions refer to the regions without annotated cells in image patches that are annotated.} are considered negative training samples. 
However, due to the complexity and large number of cells in histopathology images, completely annotating every cell of interest in the training images can be challenging. It is more practical to perform incomplete annotation, where only a fraction of the cells of interest are annotated and the unannotated areas may also contain positive instances (i.e., cells of interest)~\citep{Li2019Signet}. The annotations may even be sparse with only a few annotated cells in a training image to reduce the annotation load~\citep{bde}. Since the instances in unannotated areas are not necessarily true negative samples when the annotations are incomplete, typical network training procedures designed for complete annotations can be problematic for incomplete annotations and degrade the detection performance.

\cite{bde} propose to solve the problem of incomplete annotations for cell detection in histopathology images by calibrating the loss function during network training. Specifically, it is observed that the density of the detection boxes associated with positive instances is much greater than the box density associated with negative instances. Therefore, the \textit{Boxes Density Energy}~(BDE) is developed in \cite{bde} to calibrate the loss terms associated with the training samples in unannotated areas, where the samples with higher box density are calibrated to have smaller weights, as they are less likely to be truly negative. It is shown in~\cite{bde}, as well as in its extended journal version~\cite{BDE_journal}, that when the annotations are incomplete, the detection performance is improved with the BDE loss calibration compared with the typical training strategy that treats all instances in unannotated areas as negative. To the best of our knowledge, this is the only existing work that addresses the problem of incomplete annotations for cell detection in histopathology images\footnote{The work in~\cite{Li2019Signet} requires the annotated mask of each instance in addition to the bounding box, and thus it addresses a different problem.}, and the development of methods that may better solve this problem is still desired.

In this work, we continue to explore the problem of incomplete annotations for CNN-based cell detection in histopathology images. Since unannotated areas in incomplete annotations may include both positive and negative samples, i.e., the labels of the instances in these regions are unknown, whereas annotated samples are all positive, we propose to address the problem of incomplete annotations with a \textit{positive-unlabeled}~(PU) learning framework~\citep{Charles2008,Kiryo2017}. 
We integrate our method with advanced object detectors, where a classification loss and a box regression loss are combined for network training, and reformulate the classification loss with PU learning.
Specifically, the classification loss terms associated with negative samples are revised, so that they can be approximated with positively labeled instances and instances with unknown labels. 
We first derive the approximation for the case of binary cell detection, where only one type of cell is of interest.
Then, the approximation is extended to the case of multi-class cell detection, where more than one types of cells are to be identified given incomplete annotations.
To evaluate the proposed method, we performed experiments on four publicly available datasets of histopathology images, and for demonstration, Faster R-CNN \citep{Ren2015Faster} was used as our backbone detection network, as it has previously achieved excellent cell detection results~\citep{Sun2020,Cai2019}. 
The experimental results on the four datasets show that the proposed method leads to improved cell detection performance given incomplete annotations for training.

This manuscript is an extension of our conference paper~\citep{miccai} presented at MICCAI 2021.
In the current manuscript, we have substantially extended our work in terms of both methodology and evaluation.
Specifically, we have extended the proposed framework from the binary cell detection problem considered in~\cite{miccai} to multi-class cell detection, where the corresponding approximation of loss terms is derived and the strategy of hyperparameter selection is determined;
in addition, we have evaluated our method more comprehensively with three additional publicly available datasets under various experimental settings.
The code of the proposed method is available at \url{https://github.com/zipeizhao/PU-learning-for-cell-detection}.

We organize the remaining of the paper as follows. Section~\ref{sec:methods} presents the proposed approach to cell detection in histopathology images given incomplete annotations. In Section~\ref{sec:results}, we describe the cell detection results on the publicly available datasets. Section~\ref{sec:discussion} discusses the results and future works. Finally, Section~\ref{sec:conclusion} summarizes the proposed work.

\section{Methods}
\label{sec:methods}
In this section, we first introduce how CNN-based cell detection methods are conventionally trained given completely annotated training data. 
Then, we present the proposed approach that adapts PU learning to address the problem of incomplete annotations for cell detection in histopathology images.
Finally, the implementation details are given.

\subsection{Background: cell detection with complete annotations for training}
\label{sec:baseline}

CNN-based methods have greatly improved the performance of object detection. These methods have also been applied to cell detection and have achieved promising results.
For a typical modern CNN-based object detector, e.g., Faster R-CNN~\citep{Ren2015Faster}, convolutional layers are used to extract feature maps from input images, and the extracted feature maps are then fed into subsequent layers to predict the location and class of the objects of interest. 
Most commonly, a bounding box\footnote{Depending on the object detector, the bounding box can be defined differently. For the commonly used Faster R-CNN, it is produced by the detection network based on each anchor. For a more detailed description of the bounding box and anchor, we refer readers to~\cite{Ren2015Faster}.} $x$ is generated to indicate the position of an object of interest, which is produced by the regression head of the detector. 
For convenience, the predicted position of the bounding box $x$ is denoted by $\bm{v}=(X,Y,W,H)$, where $X$, $Y$, $W$, and $H$ represent the coordinate in the horizontal direction, coordinate in the vertical direction, width, and height of the bounding box, respectively.
The class of the object is simultaneously predicted by the classification head of the detector, where the likelihood of the instance belonging to a certain category is indicated.
For simplicity, here we discuss binary cell detection, where the detection of a specific type of cell is of interest, but its extension to multi-class cell detection---i.e., the detection of multiple types of cells---is straightforward.
In binary cell detection, the ground truth label $z$ of a bounding box $x$ is binary: $z\in\{0,1\}$, where $z=1$ represents that the bounding box contains the cell of interest, and the probability $c$ of the bounding box $x$ being positive---i.e., $z\neq 0$---predicted by the detector is between zero and one: $c\in[0,1]$.

Conventionally, to train a CNN-based cell detector, all positive instances should be annotated for the training images, and using the training data the network learns to locate and classify the cells by minimizing a loss function that sums the localization and classification errors. 
The localization loss $\mathcal{L}_{\mathrm{loc}}$ measures the difference between the predicted location $\bm{v}$ and the ground truth location $\bm{b}=(X_{b},Y_{b},W_{b},H_{b})$ of the positive training samples, where $X_{b}$, $Y_{b}$, $W_{b}$, and $H_{b}$ represent the coordinate in the horizontal direction, coordinate in the vertical direction, width, and height of the ground truth, respectively. 
A typical choice of $\mathcal{L}_{\mathrm{loc}}$ is the smooth $L_{1}$ loss function~\citep{Ren2015Faster}.
The classification loss $\mathcal{L}_{\mathrm{cls}}$ is computed from the predicted classification probability and the corresponding ground truth label as
\begin{eqnarray}
	\mathcal{L}_{\mathrm{cls}}=\frac{1}{N_{\mathrm{n}}+N_{\mathrm{p}}}\left(\sum\limits_{j=1}^{N_{\mathrm{n}}} H(c^{j}_{\mathrm{n}},0)  + \sum\limits_{i=1}^{N_{\mathrm{p}}} H(c^{i}_{\mathrm{p}},1)\right).
	\label{eq:cls}
\end{eqnarray}
Here, $i$ and $N_{\mathrm{p}}$ are the index and the total number of positive training samples (samples that have a large overlap with the labeled positive instances), respectively; $j$ and $N_\mathrm{n}$ are the index and the total number of negative training samples (samples that have no overlap with the labeled positive instances or an overlap below a threshold); $c^{i}_{\mathrm{p}}$ and $c^{j}_{\mathrm{n}}$ are the predicted classification probability for the positive samples $x^{i}_{\mathrm{p}}$ and negative samples $x^{j}_{\mathrm{n}}$, respectively; $H(\cdot,\cdot)$ measures the difference between the ground truth label and the classification result given by the network, and it is usually a cross entropy loss. With the complete annotations where every positive instance in the training images is labeled, the sum of the two loss terms $\mathcal{L}_{\mathrm{loc}}$ and $\mathcal{L}_{\mathrm{cls}}$ is minimized to learn the weights of the detection network.

\subsection{PU learning for cell detection with incomplete annotations}
\label{sec:pu}

Because there are usually a large number of cells with various appearances in histopathology images, it is challenging to annotate every positive instance. Experts may only ensure that the annotated cells are truly positive~\citep{Li2019Signet}, and the annotated cells may even appear sparse in the image to reduce the annotation load~\citep{bde}. 
In this case, the annotated training set is incomplete and only contains a subset of the positive instances.
In other words, in an incompletely annotated dataset, there are positive instances that are not annotated, and the regions with no instances labeled as positive are not necessarily all truly negative. 
Given such incomplete annotations, training the detection network with the classification loss designed for complete annotations---e.g., Eq.~(\ref{eq:cls}) for binary cell detection---is no longer accurate and could degrade the detection performance. 

Since the regions that are not labeled as positive may comprise both positive and negative samples, the instances in these regions can be considered unlabeled. This means that the incompletely annotated training dataset contains both positively labeled and unlabeled training samples. 
Therefore, to address the problem of incomplete annotations for cell detection in histopathology images, we propose to exploit PU learning, so that the classification loss that is originally computed with complete annotations can be approximated with incomplete annotations.
We first present the derivation of the approximation for the simpler case of binary cell detection. 
Then, we show how this approximation can be extended to multi-class cell detection.

\subsubsection{Binary cell detection}
\label{sec:binary}

Based on the formulation in Section~\ref{sec:baseline}, we first derive the approximation of the classification loss for binary cell detection. $\mathcal{L}_{\mathrm{cls}}$ is an approximation (empirical mean) of the expectation $\mathbb{E}_{(x,z)}[H(c,z)]$, which measures the classification inaccuracy of $c$ with respect to the ground truth label $z$, and we reformulate the computation of $\mathbb{E}_{(x,z)}[H(c,z)]$ as
\begin{eqnarray}
	&& \mathbb{E}_{(x,z)}[H(c,z)] \nonumber\\
	&=&\mathrm{Pr}(z=0) \int p(x|z=0)H(c,0)\mathrm{d}x + \mathrm{Pr}(z=1)\int p(x|z=1)H(c,1)\mathrm{d}x \nonumber\\ 
	&=&(1-\pi)\mathbb{E}_{x|z=0}[H(c,0)] +\pi \mathbb{E}_{x|z=1}[H(c,1)]
	.
	\label{eq:reform}
\end{eqnarray}
Here, we use $p(\cdot)$ to represent a probability density function, and we denote the positive class prior $\mathrm{Pr}(z=1)$ by $\pi$, which is assumed to be known. 

In incomplete annotations, positive training samples are available, whereas negative training samples cannot be determined. Therefore, the second term in Eq.~(\ref{eq:reform}) can be directly approximated with the incompletely annotated training samples but not the first term. 
However, the first term can be approximated with both positive and unlabeled training samples via PU learning~\citep{Kiryo2017}. Specifically, as $p(x)=\mathrm{Pr}(z=0)p(x|z=0) + \mathrm{Pr}(z=1)p(x|z=1)$, we have
\begin{eqnarray}
	\mathrm{Pr}(z=0)p(x|z=0) = p(x) - \mathrm{Pr}(z=1)p(x|z=1),
	\label{eq:Prp0}
\end{eqnarray}
and the first term $(1-\pi)\mathbb{E}_{x|z=0}[H(c,0)]$ in Eq.~(\ref{eq:reform}) can be rewritten as
\begin{eqnarray}
	&&(1-\pi)\mathbb{E}_{x|z=0}[H(c,0)] \nonumber\\
	&=&\mathrm{Pr}(z=0) \int p(x|z=0)H(c,0)\mathrm{d}x \nonumber\\
	&=&\int p(x) H(c,0)\mathrm{d}x - \mathrm{Pr}(z=1) \int p(x|z=1) H(c,0)\mathrm{d}x \nonumber\\
	&=& \mathbb{E}_{x}[H(c,0)] - \pi \mathbb{E}_{x|z=1}[H(c,0)].
	\label{eq:approx}
\end{eqnarray}
Then, based on Eqs.~(\ref{eq:reform}) and (\ref{eq:approx}), $\mathbb{E}_{(x,z)}[H(c,z)]$ can be rewritten as
\begin{eqnarray}
	\mathbb{E}_{(x,z)}[H(c,z)] &=& \mathbb{E}_{x}[H(c,0)] - \pi \mathbb{E}_{x|z=1}[H(c,0)] + \pi \mathbb{E}_{x|z=1}[H(c,1)].
	\label{eq:pu_exp}
\end{eqnarray}
With this derivation, the second and third terms ($\pi \mathbb{E}_{x|z=1}[H(c,0)]$ and $ \pi \mathbb{E}_{x|z=1}[H(c,1)]$, respectively) on the right hand side of Eq.~(\ref{eq:pu_exp}) can be approximated with positive training samples, and we still need to approximate the first term $\mathbb{E}_{x}[H(c,0)]$.

The original PU learning framework developed for classification problems assumes that the distribution of the unlabeled data $x_{\mathrm{u}}$ is identical to the distribution of $x$, and thus $\mathbb{E}_{x}[H(c,0)]$ can be approximated by $\mathbb{E}_{x_{\mathrm{u}}}[H(c,0)]$.
For convenience, this approximation developed for classification instead of object detection is referred to as the \textit{naive approximation} hereafter.
The naive approximation has been directly applied to object detection problems~\citep{Yang2020object}.
However, since in detection problems the unlabeled samples and positively labeled samples are drawn from the same images, the assumption that the distribution of $x_{\mathrm{u}}$ is identical to the distribution of $x$ in the naive approximation can be problematic, where some positive samples are excluded from the distribution of $x_{\mathrm{u}}$, leading to a biased approximation of $\mathbb{E}_{x}[H(c,0)]$.
To better approximate $\mathbb{E}_{x}[H(c,0)]$ for cell detection, we combine the positively labeled and unlabeled samples in the same images, and the combined samples can represent samples drawn from the distribution of $x$.
Then, $\mathbb{E}_{x}[H(c,0)]$ can be approximated as
\begin{eqnarray}
	\mathbb{E}_{x}[H(c,0)]\approx \frac{1}{N_{\mathrm{u}} + N_{\mathrm{p}}}\left(\sum_{k=1}^{N_{\mathrm{u}}}H(c_{\mathrm{u}}^{k},0) + \sum_{i=1}^{N_{\mathrm{p}}}H(c_{\mathrm{p}}^{i},0) \right).
	\label{eq:approx_unlabeled}
\end{eqnarray}
Here, $N_{\mathrm{p}}$ becomes the number of samples associated with the annotated cells in the incomplete annotations, $N_{\mathrm{u}}$ represents the number of unlabeled samples that are not associated with any annotated cells in the incomplete annotations, $k$ is the index of the unlabeled training samples, and $c^{k}_{\mathrm{u}}$ is the predicted classification probability of the $k$-th unlabeled sample~$x^{k}_{\mathrm{u}}$.
Now, all three terms on the right hand side of Eq.~(\ref{eq:pu_exp}) can be approximated with the incompletely annotated training samples.

Note that as shown in \cite{Kiryo2017}, when $(1-\pi)\mathbb{E}_{x|z=0}[H(c,0)]=\mathbb{E}_{x}[H(c,0)] - \pi \mathbb{E}_{x|z=1}[H(c,0)]$ is approximated by an expressive CNN, negative values can be produced due to overfitting. This can adversely affect the computation of $\mathbb{E}_{(x,z)}[H(c,z)]$ with Eq.~(\ref{eq:pu_exp}). Thus, 
like \cite{Kiryo2017} we use the following nonnegative approximation of $(1-\pi)\mathbb{E}_{x|z=0}[H(c,0)]$:
\begin{eqnarray}
	&&\mathbb{E}_{x}[H(c,0)] - \pi \mathbb{E}_{x|z=1}[H(c,0)] \nonumber\\
	&\approx& \max \Bigg\{0, \frac{1}{N_{\mathrm{u}} + N_{\mathrm{p}}}\Bigg(\sum_{k=1}^{N_{\mathrm{u}}}H(c_{\mathrm{u}}^{k},0) + \sum_{i=1}^{N_{\mathrm{p}}}H(c_{\mathrm{p}}^{i},0) \Bigg)  - \frac{\pi}{N_{\mathrm{p}}}\sum_{i=1}^{N_{\mathrm{p}}}H(c_{\mathrm{p}}^{i},0)\Bigg\}.
	\label{eq:approx_on_unlabeled}
\end{eqnarray}

To summarize, the derivation steps described above give us the revised classification loss $\mathcal{L}_{\mathrm{cls}}^{\mathrm{pu}}$ that approximates $\mathbb{E}_{(x,z)}[H(c,z)]$ with the PU learning framework, and it is computed as
\begin{eqnarray}
	\mathcal{L}_{\mathrm{cls}}^{\mathrm{pu}} &=&  \max \Bigg\{0, \frac{1}{N_{\mathrm{u}} + N_{\mathrm{p}}}\Bigg(\sum_{k=1}^{N_{\mathrm{u}}}H(c_{\mathrm{u}}^{k},0) + \sum_{i=1}^{N_{\mathrm{p}}}H(c_{\mathrm{p}}^{i},0) \Bigg)  - \nonumber\\
	&& \frac{\pi}{N_{\mathrm{p}}}\sum_{i=1}^{N_{\mathrm{p}}}H(c_{\mathrm{p}}^{i},0)\Bigg\} + \frac{\pi}{N_{\mathrm{p}}}\sum_{i=1}^{N_{\mathrm{p}}}H(c^{i}_{{\mathrm{p}}},1).
\end{eqnarray}
With $\mathcal{L}_{\mathrm{cls}}^{\mathrm{pu}}$, when only incomplete annotations are available for binary cell detection, the overall loss function to minimize for network training becomes  
\begin{eqnarray}
	\mathcal{L} = \mathcal{L}_{\mathrm{loc}} + \mathcal{L}_{\mathrm{cls}}^{\mathrm{pu}}.
	\label{eq:all}
\end{eqnarray}

\subsubsection{Extension to multi-class cell detection}
\label{sec:mpu}

Based on the derivation in Section~\ref{sec:binary} for binary cell detection, we further derive the approximation of the classification loss for multi-class cell detection, where the positive samples are annotated incompletely for each positive class.
Mathematically, suppose that there are $M$ classes in total, which comprise $M-1$ positive classes (cell types of interest) and one background negative class; then the ground truth label $z$ of a bounding box $x$ becomes $z\in\{0,\ldots, M-1\}$, where $z=0$ still represents the negative class and $z\in\{1,\ldots,M-1\}$ represents the positive classes.
The expectation $\mathbb{E}_{(x,z)}[H(c,z)]$ in Eq.~(\ref{eq:reform}) that is associated with the classification loss now becomes
\begin{eqnarray}
	\mathbb{E}_{(x,z)}[H(c,z)] = (1-\sum\limits_{m=1}^{M-1}\pi_{m})\mathbb{E}_{x|z=0}[H(c,0)] + \sum\limits_{m=1}^{M-1}\pi_{m} \mathbb{E}_{x|z=m}[H(c,m)],\label{eq:mpu-reform}
\end{eqnarray}
where $m\in\{1,\ldots,M-1\}$ is the positive class index and $\pi_{m}=\mathrm{Pr}(z=m)$ is the class prior (assumed to be known) for the $m$-th positive class. Note that here for multi-class detection, $c$ is a vector that comprises the predicted probabilities of all classes, and $H(\cdot,\cdot)$ computes the categorical cross entropy.
Due to the incomplete annotations, $(1-\sum_{m=1}^{M-1}\pi_{m})\mathbb{E}_{x|z=0}[H(c,0)]$ in Eq. (\ref{eq:mpu-reform}) cannot be directly approximated.

Similar to Eqs.~(\ref{eq:Prp0}) and (\ref{eq:approx}), when there are $M-1$ positive classes, because $\mathrm{Pr}(z=0)p(x|z=0)=p(x)-\sum_{m=1}^{M-1}\mathrm{Pr}(z=m) p(x|z=m)$, we have
\begin{eqnarray}
    &&(1-\sum\limits_{m=1}^{M-1}\pi_{m})\mathbb{E}_{x|z=0}[H(c,0)] \nonumber\\
 	&=&\mathrm{Pr}(z=0) \int p(x|z=0)H(c,0)\mathrm{d}x \nonumber\\
	&=&\int p(x) H(c,0)\mathrm{d}x - \sum\limits_{m=1}^{M-1}\mathrm{Pr}(z=m)\int p(x|z=m) H(c,0)\mathrm{d}x \nonumber\\
	&=& \mathbb{E}_{x}[H(c,0)] -  \sum\limits_{m=1}^{M-1}\pi_{m}\mathbb{E}_{x|z=m}[H(c,0)],
	\label{eq:mpu-approx}
\end{eqnarray}
and Eq.~(\ref{eq:mpu-reform}) becomes
\begin{eqnarray}
	\mathbb{E}_{(x,z)}[H(c,z)] = \mathbb{E}_{x}[H(c,0)]- \sum\limits_{m=1}^{M-1}\pi_{m}\mathbb{E}_{x|z=m}[H(c,0)] + \sum\limits_{m=1}^{M-1}\pi_{m} \mathbb{E}_{x|z=m}[H(c,m)].
	\label{eq:mpu_exp}
\end{eqnarray}
Like in the binary case, the first term $\mathbb{E}_{x}[H(c,0)]$ in Eq.~(\ref{eq:mpu_exp}) still needs to be determined, whereas the other terms can be computed with the labeled instances of each positive class. 

As discussed in Section~\ref{sec:binary}, the distribution of $x$ can be approximated with the combination of all positive and unlabeled samples.
Thus, we approximate $\mathbb{E}_{x}[H(c,0)]$ as
\begin{eqnarray}
	\mathbb{E}_{x}[H(c,0)] \approx \frac{1}{N_{\mathrm{u}} + \sum\limits_{m=1}^{M-1}N_{\mathrm{p}}^{m}} \left(\sum\limits_{k=1}^{N_{\mathrm{u}}}H(c_{\mathrm{u}}^{k},0) + \sum\limits_{m=1}^{M-1}\sum\limits_{i=1}^{N_{\mathrm{p}}^{m}}H(c_{\mathrm{p}}^{m,i},0)\right),
	\label{eq:mpu_approx_unlabeled}
\end{eqnarray}
where $N_{\mathrm{p}}^{m}$ represents the number of annotated samples for the $m$-th positive class and $c_{\mathrm{p}}^{m,i}$ represents the prediction probability for the $i$-th positive sample $x_{\mathrm{p}}^{m,i}$ that belongs to class~$m$.
With Eq.~(\ref{eq:mpu_approx_unlabeled}), we can approximate $\mathbb{E}_{(x,z)}[H(c,z)]$ using the incomplete annotations based on Eq.~(\ref{eq:mpu_exp}).
Note that again a nonnegative approximation of $(1-\sum_{m=1}^{M-1}\pi_{m})\mathbb{E}_{x|z=0}[H(c,0)]=\mathbb{E}_{x}[H(c,0)]-\sum_{m=1}^{M-1}\pi_{m}\mathbb{E}_{x|z=m}[H(c,0)]$ is used to avoid overfitting, which leads to
\begin{eqnarray}
&&\mathbb{E}_{x}[H(c,0)]- \sum\limits_{m=1}^{M-1}\pi_{m}\mathbb{E}_{x|z=m}[H(c,0)] \nonumber\\
&\approx& \max \Bigg\{0, \frac{1}{N_{\mathrm{u}} + \sum\limits_{m=1}^{M-1}N_{\mathrm{p}}^{m}}\left(\sum\limits_{k=1}^{N_{\mathrm{u}}}H(c_{\mathrm{u}}^{k},0) + \sum\limits_{m=1}^{M-1}\sum\limits_{i=1}^{N_{\mathrm{p}}^{m}}H(c_{\mathrm{p}}^{m,i},0)\right)  - \nonumber\\ &&\sum_{m=1}^{M-1}\frac{\pi_{m}}{N_{\mathrm{p}}^{m}}\sum_{i=1}^{N_{\mathrm{p}}^{m}}H(c_{\mathrm{p}}^{m,i},0)\Bigg\}.
\end{eqnarray}

Now, we have the classification loss $\mathcal{L}_{\mathrm{cls}}^{\mathrm{mpu}}$ for multi-class cell detection:
\begin{eqnarray}
	\mathcal{L}_{\mathrm{cls}}^{\mathrm{mpu}} &=& \max \Bigg\{0, \frac{1}{N_{\mathrm{u}} + \sum\limits_{m=1}^{M-1}N_{\mathrm{p}}^{m}}\left(\sum\limits_{k=1}^{N_{\mathrm{u}}}H(c_{\mathrm{u}}^{k},0) + \sum\limits_{m=1}^{M-1}\sum\limits_{i=1}^{N_{\mathrm{p}}^{m}}H(c_{\mathrm{p}}^{m,i},0)\right)  - \nonumber\\ 
	&&\sum\limits_{m=1}^{M-1}\frac{\pi_{m}}{N_{\mathrm{p}}^{m}}\sum_{i=1}^{N_{\mathrm{p}}^{m}}H(c_{\mathrm{p}}^{m,i},0)\Bigg\} + \sum\limits_{m=1}^{M-1}\frac{\pi_{m}}{N_{\mathrm{p}}^{m}}\sum\limits_{i=1}^{N_{\mathrm{p}}^{m}}H(c_{\mathrm{p}}^{m,i},m).\nonumber\\ 
\end{eqnarray}$\mathcal{L}_{\mathrm{cls}}^{\mathrm{mpu}}$ is used together with the localization loss (extended to multi-class detection by considering instances of all positive classes) for network training when multiple types of cells are to be detected given incomplete annotations.

\subsection{Implementation details}
\label{sec:details}

Since in detection problems it is difficult to directly estimate the class prior ($\pi$ in the binary case or $\pi_{m}$'s in the case of multi-class detection) using incompletely annotated training samples, we use a validation set, which is generally available during network training, to determine the class prior.
The detailed procedure is described below for the binary case and the multi-class case separately.

For the binary case, we consider the class prior $\pi$ a hyperparameter and search its value within a certain range. 
Because not all positive samples are labeled in incomplete annotations, the precision value computed from the validation set is no longer meaningful, and thus the value of $\pi$ is selected according to the best average recall computed from the validation set.

For the case of multi-class cell detection, there are multiple class priors $\pi_{m}$ $(m\in\{1,\ldots,M-1\})$ to be determined. A grid search for each combination of the priors does not scale with the number of classes. Therefore, we propose a more practical way of determining the class priors.
Without loss of generality, we let $\pi_{1}$ be the class prior associated with the cell type that has the largest number of annotated instances.
$\pi_{1}$ is considered a hyperparameter that is selected from a set of candidate values, and the other priors are determined from $\pi_{1}$.
More specifically, during network training, each batch is first fed into the current detector, and the number of detected cells is denoted by $N_m$ for each class $m$.
Each $\pi_{m}$ ($m\neq 1$) is updated from the fixed $\pi_{1}$ as $\pi_{m} = \pi_{1}\frac{N_{m}}{N_{1}}$, and then with the current $\pi_{m}$'s $(m\in\{1,\ldots,M-1\})$ this batch is used to compute the gradient to update the network weights.
This procedure is repeated for each batch until network training is complete.
The value of $\pi_{1}$ that achieves the best average recall on the validation set is selected. 

Our approach can be integrated with different state-of-the-art backbone detection networks that are based on the combination of localization and classification losses.
For demonstration, we selected Faster R-CNN~\citep{Ren2015Faster} (with VGG16~\citep{vgg}) as the backbone network, which is widely applied to object detection problems including cell detection~\citep{Sun2020}.
For a detailed description of Faster R-CNN, we refer readers to~\cite{Ren2015Faster}.
Intensity normalization was performed with the default normalization in Faster R-CNN, where the input image was normalized to the range of $[-1,1]$. Data augmentation was also performed according to the default operation in Faster R-CNN, where horizontal flipping was used.
The Faster R-CNN was pretrained on ImageNet~\citep{imagenet} for a better initialization of network weights. 
The Adam optimizer~\citep{adam} was used for minimizing the loss function, where the initial learning rate was set to $10^{-3}$. 
The batch size was set to 8 according to the default setting of Faster R-CNN. 
To ensure training convergence, the detection network was trained with 2580 iterations. 
The model corresponding to the last iteration was selected, as we empirically observed that model selection based on the validation set did not lead to substantially different results.


Like in Faster R-CNN, in our work the prediction and ground truth were matched based on the \textit{intersection over union} (IoU) between the anchors and ground truth boxes. Specifically, when the maximum IoU between an anchor and any ground truth box was higher than 0.7 or lower than 0.3, the anchor was considered to represent a positive or unlabeled sample, respectively; when the maximum IoU was between 0.3 and 0.7, the anchor was not used during network training.
\normalmarginpar

\section{Results}
\label{sec:results}
In this section, we present the evaluation of the proposed approach, where experiments were performed on multiple datasets under various experimental settings.
The data description and experimental settings are first given, and then the results on each dataset are described.
All experiments were performed with an NVIDIA GeForce GTX 1080 Ti GPU. 

\subsection{Data description and experimental settings}
\label{sec:data}
Four publicly available datasets developed for cell detection in histopathology images were considered to evaluate the proposed method, which are the MITOS-ATYPIA-14 dataset~\citep{Roux2013mitos}, the CRCHistoPhenotypes dataset~\citep{K2016Locality}, the TUPAC dataset~\citep{TUPAC_ori}, and the NuCLS dataset~\citep{nuclc}.
The detailed description of each dataset and the experimental settings is given below.

\subsubsection{The MITOS-ATYPIA-14 dataset}
\label{sec:mitos}

The MITOS-ATYPIA-14 dataset \citep{Roux2013mitos} aims to detect mitosis in breast cancer cells. It comprises 393 images belonging to 11 slides at $\times 40$ magnification. The slides were stained with standard \textit{Hematoxylin \& Eosin}~(H\&E) dyes, and they were imaged with an Aperio Scanscope XT scanner.
The image size is about 1539$\times $1376, and the image resolution is 0.2455 $\mu$m/pixel.
Each mitosis in this dataset was annotated with a key point by experienced pathologists, and 749 cells have been annotated.
Following~\cite{bde}, for each annotated cell we generated a 32$\times $32 bounding box centered around the key point. 
The 11 slides were split into training, validation, and test sets, and the images belonging to these slides were split accordingly for our experiment. The ratio of the number of images in the training, validation, and test sets was about 4:1:1.
We performed 5-fold cross-validation for evaluation. In each fold, the validation set was fixed, and we regrouped the training and test sets.

Due to the large image size of the MITOS-ATYPIA-14 dataset, we cropped the original images into $500\times 500$ patches, where an overlap of 100 pixels between adjacent patches in the horizontal and vertical directions was used.
To simulate incomplete annotations, like~\cite{bde} we randomly deleted the annotations in the training and validation sets until only one annotated cell per patch was kept.
Since the total number of annotated cells in the complete annotations is not large on each image patch, about 73\% of the annotated cells were kept in the training and validation sets after deletion.
Note that the deletion was performed before network training for this experiment and all other experiments as well.
Since the detection is binary for the MITOS-ATYPIA-14 dataset, network training was performed according to Section~\ref{sec:binary}.
The annotations in the test set were intact, and they were only used for the evaluation purpose.

For test images, we first detected the cells on each $500\times500$ patch, and the prediction boxes on the image patches were merged to produce the final prediction, where the coordinates of these boxes were mapped back into the image and duplicate bounding boxes were removed with \textit{non-maximum suppression}~(NMS) \citep{nms}.

The results on the MITOS-ATYPIA-14 dataset will be presented in Section~\ref{sec:result_mitos}.
First, the detection performance of the proposed method is given in Section~\ref{sec:result_mitos_performance}.
In addition, to confirm the benefit of the approximation proposed in Eq.~(\ref{eq:approx_unlabeled}) for cell detection, we have used the MITOS-ATYPIA-14 dataset to compare the proposed approximation with the naive approximation in PU learning originally developed for classification problems (described in Section~\ref{sec:binary}). The comparison of the approximation strategies will be reported in Section~\ref{sec:exp_approx}.

Moreover, we used the MITOS-ATYPIA-14 dataset to investigate the impact of detection backbones. 
Specifically, besides the VGG16 backbone~\citep{vgg}, we considered the ResNet50 and ResNet101 backbones~\citep{Resnet}, which are also commonly used for objection detection with Faster R-CNN. These backbones were integrated with the proposed method to detect cells of interest in histopathology images. The results achieved with these backbones will be reported in Section~\ref{sec:exp_bakcbone}.

Finally, in addition to the random deletion strategy described above for generating incomplete annotations, as information about the agreement on the annotations between pathologists was available in the MITOS-ATYPIA-14 dataset, we considered another scenario where pathologists choose to annotate the more confident cells. These cells are likely to be those that are easy to annotate.
Specifically, the annotated cell with the highest agreement was kept on each image patch in the training or validation set. The other experimental settings were not changed. The results achieved with this deletion strategy will be presented in Section~\ref{sec:exp_deletion}.

\subsubsection{The CRCHistoPhenotypes dataset}
\label{sec:crc}

To show that the proposed method is applicable to different datasets, we evaluated the detection performance of the proposed method on the publicly available CRCHistoPhenotypes dataset~\citep{K2016Locality}.
The CRCHistoPhenotypes dataset targets the detection of cell nuclei in colorectal adenocarcinomas. 
It comprises 100 H\&E stained images. All images have the same size of 500$\times $500 pixels, and they were cropped from non-overlapping areas of whole-slide images at a resolution of 0.55~$\mu$m/pixel. The whole-slide images were obtained with an Omnyx VL120 scanner.
A total number of 29756 nuclei were marked by experts at the center of each nucleus for detection purposes. We followed \cite{K2016Locality} and generated a 12$\times $12 bounding box centered around each annotated nucleus. 
We also performed 5-fold cross-validation for this dataset, where the images were split into training, validation, and test sets with a ratio of about 4:1:1. Like in Section~\ref{sec:mitos} the validation set was fixed, and the training and test sets were regrouped in each fold.

We cropped the images into $250\times 250$ patches with an overlap of 50 pixels horizontally and vertically between adjacent patches. 
To simulate the scenario of incomplete annotations, we considered three different cases, where the annotations were deleted at random until there was only one annotation, two annotations, or five annotations on each image patch in the training or validation set, and about 2\%, 4\%, and 9\% of the annotated cells were kept in the training and validation sets, respectively. 
Network training was performed with the incomplete annotations according to Section~\ref{sec:binary}, as only one type of cell is of interest for this dataset. 
The annotations in the test set were complete and used for evaluation only.

For test images, we generated prediction boxes for each $250\times250$ patch. These boxes were then mapped back into the image with NMS to produce the final prediction on each test image.
The detection performance of the proposed method on the CRCHistoPhenotypes dataset will be presented in Section~\ref{sec:result_crc}. 

\subsubsection{The TUPAC dataset}
\label{sec:tupac}

The TUPAC dataset~\citep{TUPAC_ori} with the alternative labels given by~\citet{TUPAC_alt} was also included for evaluation. 
We selected the first auxiliary dataset of the TUPAC dataset, which aims to detect mitosis in breast cancer. The dataset consists of H\&E images acquired at three centers, and we used the 23 cases from the first center. 
The 23 cases were split into training, validation, and test sets with a ratio of about 4:1:1.
Each case is associated with an image, the size of which is about $20000\times20000$. The images were acquired on an Aperio ScanScope XT scanner at $\times 40$ magnification with a resolution of 0.25~$\mu$m/pixel.

Due to the large image size of this dataset, we cropped the images into $500\times 500$ patches (without overlap), and patches without cells of interest were discarded.
The dataset provides both complete and incomplete annotations for the images.
However, the difference in the number of annotated cells between the complete and incomplete annotations is small (1359 vs 1273 for the 23 cases).
Therefore, based on the original incomplete annotations, we further randomly deleted the annotations in the training and validation sets until there was only one annotated cell per patch. This led to new incomplete annotations that comprised about 63\% of all annotated cells in the training and validation sets.
Like for the MITOS-ATYPIA-14 dataset, for each annotated mitosis we generated a $32\times 32$ bounding box centered around it, and network training was performed with the new incomplete annotations according to Section~\ref{sec:binary}.
The annotations in the test set were complete and only used for evaluation.

Since the size of the original image is large, evaluation was performed directly on the image patches.
The detection performance of the proposed method on the TUPAC dataset will be presented in Section~\ref{sec:result_tupac}.

\subsubsection{The NuCLS dataset}
\label{sec:nuclc}

To demonstrate the applicability of the proposed method to multi-class cell detection, we performed experiments on the NuCLS dataset \citep{nuclc}. The dataset provides labeled nuclei of seven classes of cells in breast cancer images from \textit{The Cancer Genome Atlas}~(TCGA)~\citep{TCGA}. 
Note that the NuCLS dataset provides annotations~(bounding boxes) of different quality, and we used the subset of the images associated with the high-quality cell annotations for evaluation, where initial annotations have been manually corrected by study coordinators based on the feedback from a senior pathologist. 
This subset comprises 1744 images, and the image size is about 400 $\times$ 400. 
The images are from the scanned diagnostic H\&E slides (mostly at $\times$20--40 magnification) generated by the TCGA Research Network and were accessed with the Digital Slide Archive repository.
Since not all seven cell types have a large number of annotated instances, for our experiment, we selected three types of cells for which a large number of annotations were made on these images, and they are the tumor class (with 21067 annotated nuclei), the lymphocyte class (with 13630 annotated nuclei), and the stromal class (with 9132 annotated nuclei).

The images were split into training, validation, and test sets with a ratio of about 4:1:1, and they were directly used for training and testing without cropping. To simulate the scenario of incomplete annotations, for each cell type, if there were more than ten annotated instances in an image in the training or validation set, we randomly deleted the annotations until only ten annotations remained. 
After deletion 50\%, 60\%, and 71\% of the annotated tumor, lymphocyte, and stromal cells were kept in the training and validation sets, respectively.
Since multiple types of cells were of interest here, network training was performed with the incomplete annotations according to Section~\ref{sec:mpu}. 
The annotations in the test set were complete, and they were only used for evaluation. 
The detection performance of the proposed method on the NuCLS dataset will be presented in Section~\ref{sec:result_nucls}.

\subsubsection{Competing methods and upper bound performance}

In the experiment, the proposed method was compared with two competing methods, which, for fair comparison, used the same backbone Faster R-CNN detection network. 
The first one is the baseline Faster R-CNN model (also pretrained on the ImageNet dataset as described in Section~\ref{sec:details}), which neglected that the annotations were incomplete and simply considered the unlabeled regions truly negative. Note that here the baseline Faster R-CNN was trained with the standard cross entropy loss. Although it is also possible to use other losses that address imbalanced samples, such as the weighted cross entropy loss or focal loss~\citep{Lin2017RetinaNet}, we have empirically observed that they led to worse performance, and thus they were not considered.\footnote{The performance achieved with these alternative losses is discussed in Appendix~\ref{app:loss}.}
The second one is the BDE method~\citep{bde,BDE_journal} that addresses the problem of incomplete annotations for cell detection with a calibrated loss function, and it was integrated with the Faster R-CNN architecture with network weights initialized on ImageNet. 

In addition to these competing methods, we have also computed the upper bound performance that was achieved with the complete annotations for training. Specifically, the original complete annotations without deletion were used in the training and validation sets, and Faster R-CNN was trained with these training and validation sets with the standard training procedure, as no PU learning was needed for complete annotations.

\begin{figure}[!t]

	\centerline{\includegraphics[width=0.99\textwidth]{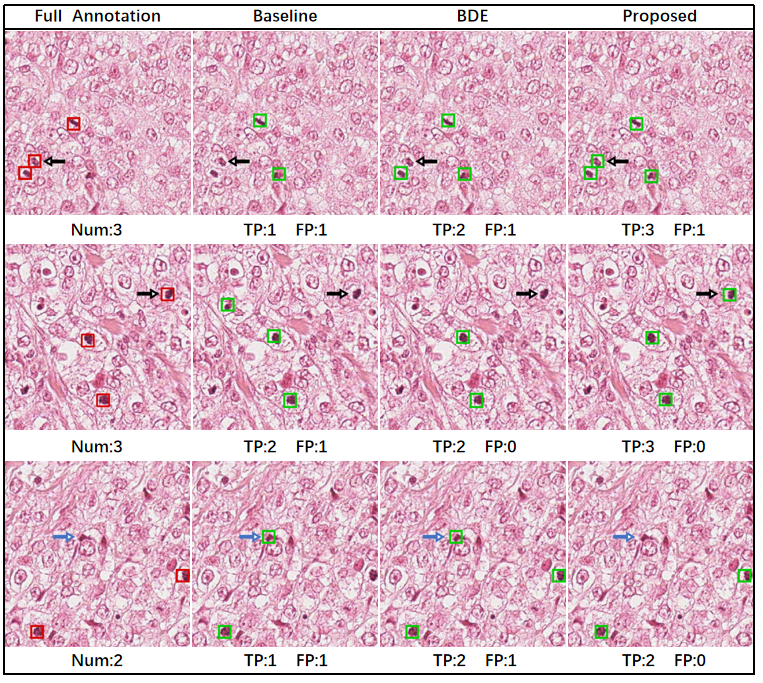}}
	\caption{Examples of representative detection results on test patches for the MITOS-ATYPIA-14 dataset. The gold standard full annotations and the numbers of annotated cells in the full annotations are also shown for reference. TP and FP represent the numbers of true positive and false positive detection results on the patch, respectively. Note the regions highlighted by arrows for comparison. The black arrows indicate examples of true positive cases given by the proposed method but missed by the competing methods, whereas the blue arrows indicate examples of true negative cases given by the proposed method but labeled as positive by BDE.} \label{fig:mitos}
\end{figure}

\subsection{Results on the MITOS-ATYPIA-14 dataset}
\label{sec:result_mitos}

\subsubsection{Detection performance}
\label{sec:result_mitos_performance}

We first present the detection results of the proposed method on the MITOS-ATYPIA-14 dataset when incomplete annotations were obtained with random deletion. As described in Section~\ref{sec:details}, the class prior $\pi$ was determined based on the validation set. The candidate values of $\pi$ ranged from 0.025 to 0.050 with an increment of 0.005. 
Note that $\pi$ was selected for each fold independently, and the selected value was consistent (0.035 to 0.045) across the folds.\footnote{A detailed analysis of the sensitivity to the class prior $\pi$ is given in Appendix~\ref{app:pi}.}

Examples of the detection results of the proposed and competing methods are shown in Fig.~\ref{fig:mitos}, where the bounding boxes predicted by each method on representative test patches are displayed. For reference, the gold standard full annotations on the test patches are also shown, and the numbers of true positive and false positive detection results on a patch are indicated for each method.
In these cases, our method compares favorably with the competing methods by either producing more true positive boxes without increasing the number of false positive boxes or reducing the number of false positive boxes with preserved true positive boxes.

For quantitative evaluation, we computed the average recall, average precision, and average F1-score of the detection results on the test set for each method and each fold, as well as the upper bound performance achieved with the complete annotation for training, and they are shown in Tables~\ref{tab:mitos_fold} and \ref{tab:mitos_f1_fold}.\footnote{To investigate the impact of random effects, we further repeated the experiment with multiple independent runs for the proposed and competing methods, and these results are reported in Appendix~\ref{app:random}.}
Compared with the competing methods, the proposed method has higher recall, precision, and F1-score, which indicate the better detection accuracy of our method, and this improvement is consistent across the folds.
In addition, the F1-score of our method is closer to the upper bound than the competing methods.
We also computed the means and standard deviations of the average recall, average precision, and average F1-score of the five folds, and compared the proposed method with the competing methods using paired Student's $t$-tests. These results are shown in Table~\ref{tab:mitos_m_std_p}. Consistent with Tables~\ref{tab:mitos_fold} and \ref{tab:mitos_f1_fold}, the proposed method has higher recall, precision, and F1-score, and the improvement of our method is statistically significant.

\begin{table*}[!t]
	\centering
	\caption{The average recall and average precision of the detection results on the test set for each fold for the MITOS-ATYPIA-14 dataset when incomplete annotations were obtained with random deletion. The best results are highlighted in bold. The upper bound performance is also shown for reference.}
	\label{tab:mitos_fold}
	\resizebox{0.99\textwidth}{!}{
		\begin{tabular}{c >{\centering\arraybackslash}p{1.2cm} >{\centering\arraybackslash}p{1.6cm} >{\centering\arraybackslash}p{1.2cm} >{\centering\arraybackslash}p{1.6cm} >{\centering\arraybackslash}p{1.2cm} >{\centering\arraybackslash}p{1.6cm} >{\centering\arraybackslash}p{1.2cm} >{\centering\arraybackslash}p{1.6cm} >{\centering\arraybackslash}p{1.2cm} >{\centering\arraybackslash}p{1.6cm}}
			\toprule
			\multirow{2}{*}{Method} & \multicolumn{2}{c  }{Fold 1}& \multicolumn{2}{c  }{Fold 2}&\multicolumn{2}{c }{Fold 3}&\multicolumn{2}{c  }{Fold 4}& \multicolumn{2}{c }{Fold 5} \\
			\cmidrule(r){2-3}
			\cmidrule(r){4-5}
			\cmidrule(r){6-7}
			\cmidrule(r){8-9}
			\cmidrule(r){10-11}
			&    Recall & Precision        &      Recall &  Precision     &   Recall  & Precision       &     Recall  &  Precision    &   Recall  &  Precision     \\
			\hline
			\hline
			Baseline 				&    0.602  		&  0.412          	&      0.504  			&   0.401       		&   0.642  &   0.357        &      0.460  &   0.372      &    0.643  &  0.473   \cr
			BDE      					&    0.634  		&  0.438          	&      0.532  			&   0.421       		&    0.659  &   0.368        &      0.482  &    0.418     &    0.682  &  0.489   \cr
			Proposed 				&    \bf{0.645}	&  \bf{0.441}		&      \bf{0.538}  	&   \bf{0.445}    &    \bf{0.667}  &   \bf{0.381}  & \bf{0.492}  & \bf{0.429}  & \bf{0.698} & \bf{0.501} \cr
			\hline
			Upper Bound 				&    0.639  		&  0.479          	&      0.552  			&   0.456       		&   0.678  &   0.389        &      0.502  &   0.441      &    0.692  &  0.541
			\\
			\bottomrule
	\end{tabular}}
\end{table*}

\begin{table*}[!t]
	
	\centering
	\caption{The average F1-score of the detection results on the test set for each fold for the MITOS-ATYPIA-14 dataset when incomplete annotations were obtained with random deletion. The best results are highlighted in bold. The upper bound performance is also shown for reference.}
	\label{tab:mitos_f1_fold}
	\resizebox{0.65\textwidth}{!}{
		\begin{tabular}{c >{\centering\arraybackslash}p{1.5cm} >{\centering\arraybackslash}p{1.5cm} >{\centering\arraybackslash}p{1.5cm} >{\centering\arraybackslash}p{1.5cm} >{\centering\arraybackslash}p{1.5cm} }
			\toprule
			\multirow{2}{*}{Method}& \multicolumn{5}{c }{F1-score} \\
			\cmidrule(r){2-6} 
			&    Fold 1 & Fold 2        &      Fold 3 &  Fold 4     &   Fold 5            \\
			\hline
			\hline
			Baseline 				&    0.490  		&  0.446          	&      0.458  			&   0.411       		&    0.545     \cr
			BDE      					&    0.519  		&  0.470          	&      0.472  			&   0.448       		&    0.570     \cr
			Proposed 				&    \bf{0.524}	&  \bf{0.487}		&      \bf{0.484}  	&   \bf{0.457}    &    \bf{0.583} \cr
			\hline
			Upper Bound      					&    0.546  		&  0.499          	&      0.494  			&   0.469       		&    0.607
			\\
						\bottomrule
	\end{tabular}}
\end{table*}

\begin{table}[!t]
	
	\centering
	\caption{The means and \textit{standard deviations}~(stds) of the average recall, average precision, and average F1-score of the five folds for the MITOS-ATYPIA-14 dataset when incomplete annotations were obtained with random deletion. The best results are highlighted in bold. The upper bound performance is also shown for reference. Asterisks indicate that the difference between the proposed method and the competing method is significant using a paired Student's $t$-test after Benjamini-Hochberg correction for multiple comparisons. ($^{*}p<0.05$, $^{**}p<0.01$)}
	\label{tab:mitos_m_std_p}
	\resizebox{0.99\textwidth}{!}{
		 \begin{tabular}{c >{\centering\arraybackslash}p{1.3cm} >{\centering\arraybackslash}p{1.4cm} >{\centering\arraybackslash}p{1.5cm} >{\centering\arraybackslash}p{1.3cm} >{\centering\arraybackslash}p{1.4cm} >{\centering\arraybackslash}p{1.5cm} >{\centering\arraybackslash}p{1.3cm} >{\centering\arraybackslash}p{1.4cm} >{\centering\arraybackslash}p{1.5cm}}
			\toprule
			\multirow{2}{*}{Method} & \multicolumn{3}{c }{Recall}& \multicolumn{3}{c }{Precision}&\multicolumn{3}{c }{F1-score} \\
			\cmidrule(r){2-4} \cmidrule(r){5-7}\cmidrule(r){8-10}
			& mean &  std &  $p$  & mean  & std  & $p$  & mean  & std  & $p$      \\
			\hline
			\hline
			Baseline & 0.570 & 0.075   & **   & 0.403   & 0.040 & ** &0.470&0.045&** \\ 
			BDE & 0.598  & 0.077 & **  & 0.427  & 0.039 & *& 0.496  & 0.044& ** \\
			 Proposed  & \textbf{0.608}  & \textbf{0.079} & -  & \textbf{0.439}  & \textbf{0.038} & -& \textbf{0.507}  & \textbf{0.044}& -\\
			 \hline
			 Upper Bound& 0.613  & 0.074 & -  & 0.461  & 0.049 & -& 0.523  & 0.048& -
			\\
			\bottomrule
	\end{tabular}}
\end{table}

\subsubsection{Comparison with the naive approximation}
\label{sec:exp_approx}



We then performed experiments to confirm the benefit of the approximation developed in Eq.~(\ref{eq:approx_unlabeled}) for detection problems.
As described in Section~\ref{sec:mitos}, cell detection with incompletely annotated training data was performed with the naive approximation used in PU learning for classification problems on the MITOS-ATYPIA-14 dataset.

\begin{table}[!t]
	\centering
	\caption{The average recall, average precision, and average F1-score of the detection results achieved with the naive approximation on the test set for each fold for the MITOS-ATYPIA-14 dataset. (Incomplete annotations were obtained with random deletion.) The means and \textit{standard deviations} (stds) of these results are also shown.}
	\label{tab:mitos_pu_exp}
	\resizebox{0.8\columnwidth}{!}{
		\begin{tabular}{ >{\centering\arraybackslash}p{1.5cm} >{\centering\arraybackslash}p{1.2cm} >{\centering\arraybackslash}p{1.2cm} >{\centering\arraybackslash}p{1.2cm} >{\centering\arraybackslash}p{1.2cm} >{\centering\arraybackslash}p{1.2cm}
		>{\centering\arraybackslash}p{1.2cm} >{\centering\arraybackslash}p{1.2cm}}
			\toprule[1pt]
			& Fold 1 & Fold 2        &      Fold 3 &  Fold 4     &   Fold 5    & mean & std        \\
			\hline
			\hline
			Recall &0.656 	& 0.532 		& 0.681       	&  0.487     &  0.715  & 0.614 & 0.089\\

            Precision & 0.423	&  	0.435	&  0.370      	&   0.420    & 0.484  & 0.426  &0.036\\
    
			F1-score & 0.514	&  	0.477	&   0.479     	&    0.451   & 0.577  & 0.500& 0.044\\
									
			\bottomrule[1pt]
	\end{tabular}}
\end{table}

The average recall, average precision, and average F1-score of each fold achieved with the naive approximation are listed in Table~\ref{tab:mitos_pu_exp}, as well as their mean values and standard deviations.
By comparing Table~\ref{tab:mitos_pu_exp} with Table~\ref{tab:mitos_f1_fold}, we can see that for each fold the F1-score of the naive approximation is worse than the result of the proposed method, and it is even worse than the BDE result for the first fold. These results indicate the benefit of the proposed approximation.

\subsubsection{Impact of different backbones}
\label{sec:exp_bakcbone}

Next, we investigated the applicability of the proposed method to different detection backbones with the MITOS-ATYPIA-14 dataset as described in Section~\ref{sec:mitos}, where ResNet50 and ResNet101~\citep{Resnet} were considered for Faster R-CNN. 
The competing methods were also integrated with these backbones, and they were compared with the proposed method.

The results are summarized in Table~\ref{tab:backbone_m_std_p} (together with the upper bound computed with the different backbones), where the means and standard deviations of the average recall, average precision, and average F1-score of the five folds are listed. The proposed method is also compared with the competing methods using paired Student's $t$-tests in Table~\ref{tab:backbone_m_std_p}.
With these different backbones, the proposed method still has higher recall, precision, and F1-score than the competing methods, and the improvement is statistically significant.

\begin{table*}[!t]
	\centering
	\caption{The means and \textit{standard deviations} (stds) of the average recall, average precision, and average F1-score of the five folds for the MITOS-ATYPIA-14 dataset with different detection backbones. (Incomplete annotations were obtained with random deletion.) The best results are highlighted in bold. The upper bound performance is also shown for reference. Asterisks indicate that the difference between the proposed method and the competing method is significant using a paired Student's $t$-test after Benjamini-Hochberg correction for multiple comparisons. ($^{*}p<0.05$, $^{**}p<0.01$)}
	\label{tab:backbone_m_std_p}
	\resizebox{0.99\textwidth}{!}{
		\begin{tabular}{
		c >{\centering\arraybackslash}p{2.8cm} >{\centering\arraybackslash}p{1.5cm} >{\centering\arraybackslash}p{1.5cm} >{\centering\arraybackslash}p{1.5cm} >{\centering\arraybackslash}p{1.3cm} >{\centering\arraybackslash}p{1.5cm} >{\centering\arraybackslash}p{1.5cm} >{\centering\arraybackslash}p{1.3cm} >{\centering\arraybackslash}p{1.5cm} >{\centering\arraybackslash}p{1.5cm}}
			\toprule
			\multirow{2}{*}{Backbone}& \multirow{2}{*}{Method} & \multicolumn{3}{c }{Recall}& \multicolumn{3}{c }{Precision}& \multicolumn{3}{c }{F1-score} \\
			\cmidrule(r){3-5} \cmidrule(r){6-8}\cmidrule(r){9-11}
			& & mean &  std &  $p$  & mean  & std  & $p$  & mean  & std  & $p$      \\
			\hline
			\hline
			\multirow{4}{*}{ResNet50}&Baseline & 0.571 & 0.067   & **   & 0.396   & 0.047 & ** & 0.463   & 0.051 & ** \\ 
			&BDE &0.601 &0.071 &** &0.422 &0.051 &* &0.495 &0.054 &**\\
			&Proposed & \textbf{0.619} & \textbf{0.072}  & - & \textbf{0.441}  & \textbf{0.049} & -& \textbf{0.513}  & \textbf{0.055} & -\\
			\cmidrule(r){2-11}
			&Upper Bound & 0.627& 0.070& -&0.458& 0.062& -&0.526& 0.061 & -\\
			\hline
			\multirow{4}{*}{ResNet101}&Baseline & 0.580 & 0.070  & **   & 0.379   & 0.043 & ** & 0.459   & 0.046& ** \\ 
			&BDE &0.611 &0.072 &** &0.391 &0.044 &* &0.478 &0.045 &*\\
			&Proposed & \textbf{0.632} & \textbf{0.071}  & -& \textbf{0.403}  & \textbf{0.042} & -& \textbf{0.490}  & \textbf{0.045} & -\\
			\cmidrule(r){2-11}
			&Upper Bound &0.636 &0.070 &- &0.440 &0.052 &- &0.517 &0.052 &-\\
			\bottomrule
	\end{tabular}}
\end{table*}

\subsubsection{Impact of annotation strategies}
\label{sec:exp_deletion}

We then performed experiments with the other annotation strategy, where incomplete annotations were generated based on the agreement on the annotations between pathologists as described in Section~\ref{sec:mitos}.
The quantitative results are summarized in Table~\ref{tab:mitos_m_std_p_ag}, where the means and standard deviations of the average recall, average precision, and average F1-score of the five folds are listed.
The proposed method is also compared with the competing methods using paired Student's $t$-tests.
The proposed method still has higher recall, precision, and F1-score than the two competing approaches, and the improvement is significant.


\begin{table}[!t]
	\centering
	\caption{
	The means and \textit{standard deviations}~(stds) of the average recall, average precision, and average F1-score of the five folds for the MITOS-ATYPIA-14 dataset when incomplete annotations were obtained based on the agreement between pathologists. The best results are highlighted in bold. Asterisks indicate that the difference between the proposed method and the competing method is significant using a paired Student's $t$-test after Benjamini-Hochberg correction for multiple comparisons. ($^{**}p<0.01$)}
	\label{tab:mitos_m_std_p_ag}
	\resizebox{0.99\textwidth}{!}{
		 \begin{tabular}{c >{\centering\arraybackslash}p{1.3cm} >{\centering\arraybackslash}p{1.4cm} >{\centering\arraybackslash}p{1.5cm} >{\centering\arraybackslash}p{1.3cm} >{\centering\arraybackslash}p{1.4cm} >{\centering\arraybackslash}p{1.5cm} >{\centering\arraybackslash}p{1.3cm} >{\centering\arraybackslash}p{1.4cm} >{\centering\arraybackslash}p{1.5cm}}
			\toprule
			\multirow{2}{*}{Method} & \multicolumn{3}{c }{Recall}& \multicolumn{3}{c }{Precision}&\multicolumn{3}{c }{F1-score} \\
			\cmidrule(r){2-4} \cmidrule(r){5-7}\cmidrule(r){8-10}
			& mean &  std &  $p$  & mean  & std  & $p$  & mean  & std  & $p$      \\
			\hline
			\hline
			Baseline & 0.575 & 0.072   & **   & 0.404   & 0.041 & ** &0.473&0.045&** \\ 
			BDE& 0.598  & 0.077 & **  & 0.427  & 0.041 & **& 0.497  & 0.044& ** \\
			 Proposed  & \textbf{0.608}  & \textbf{0.077} & -  & \textbf{0.443}  & \textbf{0.041} & -& \textbf{0.512}  & \textbf{0.045}& -\\
			
			\bottomrule
	\end{tabular}}
\end{table}

\subsection{Results on the CRCHistoPhenotypes dataset}
\label{sec:result_crc}

\begin{figure*}[!t]

	\centerline{\includegraphics[width=0.9\textwidth]{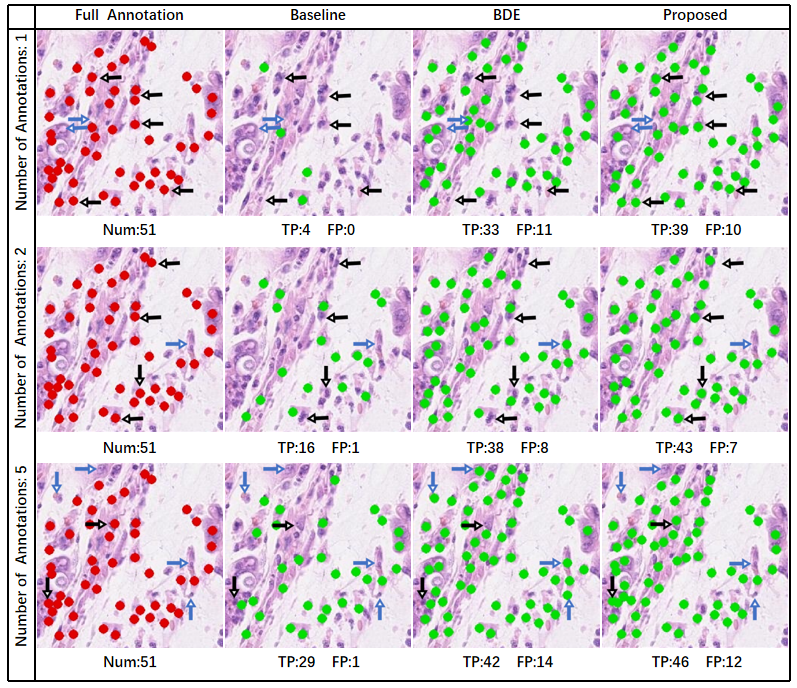}}
	\caption{Examples of representative detection results on a test patch for the CRCHistoPhenotypes dataset when one, two, or five annotated cells were available in each training patch. The gold standard full annotations and the numbers of annotated cells in the full annotations are also shown for the test patch for reference. TP and FP represent the numbers of true positive and false positive detection results on the patch, respectively. Note the regions highlighted by arrows for comparison. The black arrows indicate examples of true positive cases given by the proposed method but missed by the competing methods, whereas the blue arrows indicate examples of true negative cases given by the proposed method but labeled as positive by BDE. Note that the same test patch is used for the different numbers of annotated cells for training, but the highlighted regions are different for these cases.} 
	\label{fig:CRC}
\end{figure*}

We further present the detection results on the CRCHistoPhenotypes dataset. As described in Section~\ref{sec:crc}, different cases were considered, where the number of annotated cells in the incompletely annotated dataset varied (one, two, and five per patch, respectively). 
The candidate values of the class prior $\pi$ ranged from 0.1 to 0.4 with an increment of 0.05, and for each case of annotated cells, the selected value (0.3 or 0.35) based on the validation set was consistent across the folds.

Examples of the detection results of each method on a representative test patch are shown in Fig.~\ref{fig:CRC} for the different cases of annotated cells for training.
The complete annotations on the test patch are also shown for reference.
Note that since there are a large number of instances in the patch, for the visualization purpose, only the centers of the bounding boxes are shown in Fig.~\ref{fig:CRC} as dots. 
The numbers of true positive and false positive detection results on the patch are indicated for each method and each case.
The baseline method only detected a very small fraction of the cells of interest, which are much fewer than the results of BDE and the proposed method.
Compared with the BDE results, the detection results given by the proposed method better match the gold standard full annotations, and the proposed method produced more true positives and fewer false positives for the examples.

Quantitatively, for each fold we computed the average recall, average precision, and average F1-score of the detection results on the test set, and we also computed the corresponding upper bound performance. These results are shown in Tables~\ref{tab:CRC_fold} and \ref{tab:crc_f1_fold} for each method and each case of annotated cells.
In all cases, the proposed method has higher recall, precision, and F1-score than the two competing approaches, and the results of our method are closer to the upper bound.
The means and standard deviations of the average recall, average precision, and average F1-score of the five folds were also computed and are summarized in Table~\ref{tab:CRC_m_std_p}, where the proposed method is compared with the competing methods using paired Student's $t$-tests. In most cases, the proposed method statistically significantly outperforms the competing methods.
\begin{table*}[!t]
	\centering
	\arrayrulecolor{black}
	\caption{The average recall and average precision of the detection results on the test set for each fold for the CRCHistoPhenotypes dataset when different numbers of annotated cells were available in each training patch. The best results are highlighted in bold. The upper bound performance is also shown for reference.}
	\label{tab:CRC_fold}
	\resizebox{0.99\textwidth}{!}{
		\begin{tabular}{c >{\centering\arraybackslash}p{2.8cm}
				>{\centering\arraybackslash}p{1.0cm} >{\centering\arraybackslash}p{1.6cm} >{\centering\arraybackslash}p{1.0cm} >{\centering\arraybackslash}p{1.6cm} >{\centering\arraybackslash}p{1.0cm} >{\centering\arraybackslash}p{1.6cm} >{\centering\arraybackslash}p{1.0cm} >{\centering\arraybackslash}p{1.6cm} >{\centering\arraybackslash}p{1.0cm} >{\centering\arraybackslash}p{1.6cm}}
			\toprule
			\multirow{1}{*}{Number of}&\multirow{2}{*}{Method} & \multicolumn{2}{c }{Fold 1}& \multicolumn{2}{c }{Fold 2}&\multicolumn{2}{c }{Fold 3}&\multicolumn{2}{c }{Fold 4}& \multicolumn{2}{c}{Fold 5} \\
            \cmidrule(r){3-4}
            \cmidrule(r){5-6}
            \cmidrule(r){7-8}
            \cmidrule(r){9-10}
            \cmidrule(r){11-12}
			Annotations& &    Recall & Precision        &      Recall &  Precision     &   Recall  & Precision       &     Recall  &  Precision    &   Recall  &  Precision     \\
			\hline
			\hline
			\multirow{3}{*}{1}& Baseline 				&    0.112  		&  0.101          	&      0.115  			&   0.098       		&    0.103  &   0.096        &      0.098  &   0.091      &    0.112  &  0.091   \cr
			& BDE      					&    0.524  		&  0.462          	&      0.501  			&   0.421       		&    0.574  &   \bf{0.471}        &      0.620  &    0.554     &    0.521  &  0.464   \cr
			& Proposed 				&    \bf{0.545}	&  \bf{0.470}		&      \bf{0.530}  	&   \bf{0.463}    &    \bf{0.599}  &   \bf{0.471}  & \bf{0.625}  & \bf{0.559}  & \bf{0.537} & \bf{0.473}    \\
			\hline
			\multirow{3}{*}{2}& Baseline 				&    0.120  		&  0.105          	&      0.131  			&   0.101       		&    0.092  &   0.101        &      0.147  &   0.091      &    0.146  &  0.098   \cr
			& BDE      		 &    0.534  &  0.475			&    0.527  		&  0.462          	&      0.574  			&   0.473       		&    0.606  &   0.551        &      0.536  &    0.474       \cr
			& Proposed 				&    \bf{0.557}	&  \bf{0.484}		&      \bf{0.554}  	&   \bf{0.478}    &    \bf{0.605}  &   \bf{0.539}  & \bf{0.628}  & \bf{0.557}  & \bf{0.551} & \bf{0.487}    \\
			\hline
			\multirow{3}{*}{5}& Baseline 				&    0.204  		&  0.198          	&      0.213  			&   0.241       		&    0.265  &   0.256        &      0.274  &   0.260      &    0.208  &  0.253   \cr
			& BDE      					&    0.556  		&  0.483          	&      0.577  			&   0.471       		&    0.631  &   0.522        &      0.654  &    0.560     &    0.581  &  0.485   \cr
			& Proposed 				&    \bf{0.562}	&  \bf{0.488}		&      \bf{0.598}  	&   \bf{0.486}    &    \bf{0.638}  &   \bf{0.532}  & \bf{0.667}  & \bf{0.566}  & \bf{0.592} & \bf{0.489}  \cr
			\hline
			\multirow{1}{*}{-}&Upper Bound &    0.671	&  0.601		&      0.682  	&   0.612    &    0.696  &   0.629  & 0.701  & 0.639  & 0.689 & 0.605\\
			\bottomrule
	\end{tabular}}
\end{table*}

\begin{table*}[!t]
	\centering
	\caption{The average F1-score of the detection results on the test set for each fold for the CRCHistoPhenotypes dataset when different numbers of annotated cells were available in each training patch. The best results are highlighted in bold. The upper bound performance is also shown for reference.}
	\label{tab:crc_f1_fold}
	\resizebox{0.8\textwidth}{!}{
		\begin{tabular}{c >{\centering\arraybackslash}p{2.8cm} >{\centering\arraybackslash}p{1.4cm} >{\centering\arraybackslash}p{1.4cm} >{\centering\arraybackslash}p{1.4cm} >{\centering\arraybackslash}p{1.4cm} >{\centering\arraybackslash}p{1.4cm} }
			\toprule
			\multirow{1}{*}{Number of}&\multirow{2}{*}{Method}& \multicolumn{5}{c }{F1-score} \\
			\cmidrule(r){3-7} 
			Annotations& &    Fold 1 & Fold 2        &      Fold 3 &  Fold 4     &   Fold 5            \\
			\hline
			\hline
			\multirow{3}{*}{1}&Baseline 				&    0.101  		&  0.112          	&      0.105  			&   0.095       		&    0.096     \cr
			&BDE      					&    0.489  		&  0.451          	&      0.521  			&   0.582       		&    0.483     \cr
			&Proposed 				&    \bf{0.501}	&  \bf{0.490}		&      \bf{0.530}  	&   \bf{0.591}    &    \bf{0.497}\\
			\hline
			\multirow{3}{*}{2}&Baseline 				&    0.115  		&  0.116          	&      0.103  			&   0.108       		&    0.112     \cr
			&BDE      					&    0.507  		&  0.487          	&      0.512  			&   0.571       		&    0.502     \cr
			&Proposed 				&    \bf{0.513}	&  \bf{0.508}		&      \bf{0.564}  	&   \bf{0.596}    &    \bf{0.519}\\
			\hline
			\multirow{3}{*}{5}&Baseline 				&    0.205  		&  0.221          	&      0.257  			&   0.269       		&    0.228     \cr
			&BDE      					&    0.517  		&  0.519          	&      0.581  			&   0.606       		&    0.523     \cr
			&Proposed 				&    \bf{0.524}	&  \bf{0.532}		&      \bf{0.587}  	&   \bf{0.614}    &    \bf{0.530} \cr
			\hline
			\multirow{1}{*}{\textcolor{red}{-}}& 	Upper Bound			&    0.634  		&  0.645         	&      0.661  			&   0.669       		&    0.644\\	
			
			\bottomrule
	\end{tabular}}
\end{table*}

\begin{table*}[!t]
	\centering
	\caption{The means and \textit{standard deviations}~(stds) of the average recall, average precision, and average F1-score of the five folds for the CRCHistoPhenotypes dataset when different numbers of annotated cells were available in each training patch. The best results are highlighted in bold. The upper bound performance is also shown for reference. Asterisks indicate that the difference between the proposed method and the competing method is significant using a paired Student's $t$-test after Benjamini-Hochberg correction for multiple comparisons. ($^{*}p<0.05$, $^{**}p<0.01$, $^{***}p<0.001$, n.s. $p>0.05$)}
	\label{tab:CRC_m_std_p}
	\resizebox{0.99\textwidth}{!}{
		\begin{tabular}{
		c >{\centering\arraybackslash}p{2.8cm} >{\centering\arraybackslash}p{1.3cm} >{\centering\arraybackslash}p{1.0cm} >{\centering\arraybackslash}p{1.5cm} >{\centering\arraybackslash}p{1.3cm} >{\centering\arraybackslash}p{1.0cm} >{\centering\arraybackslash}p{1.5cm} >{\centering\arraybackslash}p{1.3cm} >{\centering\arraybackslash}p{1.0cm} >{\centering\arraybackslash}p{1.5cm}}
			\toprule
			\multicolumn{1}{c }{ Number of}&\multirow{2}{*}{Method} & \multicolumn{3}{c }{Recall}& \multicolumn{3}{c }{Precision}& \multicolumn{3}{c }{F1-score} \\
			\cmidrule(r){3-5} \cmidrule(r){6-8}\cmidrule(r){9-11}
			Annotations& & mean &  std &  $p$  & mean  & std  & $p$  & mean  & std  & $p$      \\
			\hline
			\hline
			\multirow{3}{*}{1}&Baseline &0.108  & 0.044   & ***   & 0.095   & 0.047 & *** &  0.102  & 0.006 & *** \\ 
			&BDE & 0.548  & 0.055 & *  &  0.474 & 0.052 & n.s.&  0.505 & 0.044 & * \\
			&Proposed & \textbf{0.567} & \textbf{0.060}  & - & \textbf{0.487}  & \textbf{0.048} & -& \textbf{0.522}  & \textbf{0.037} & -\\
			\hline
			\multirow{3}{*}{2}&Baseline &0.127&0.020&***&0.099&0.004&*** &0.111&0.005&*** \\ 
			&BDE &0.556&0.030&**&0.487&0.032&n.s.&0.515&0.029&* \\
			&Proposed & \textbf{0.579} & \textbf{0.031}  & - & \textbf{0.509}  & \textbf{0.032} & -& \textbf{0.540}  & \textbf{0.034} & -\\
			\hline
			\multirow{3}{*}{5}&Baseline &0.233&0.030&***&0.242&0.023&***&0.236&0.024&***  \\ 
			&BDE & 0.600&0.037&*&0.504&0.033&* &0.549&0.037&**\\
			&Proposed & \textbf{0.611} & \textbf{0.037}  & - & \textbf{0.512} & \textbf{0.032} & -& \textbf{0.558}  & \textbf{0.036} & -\\
			\hline
			-&Upper Bound & 0.688 & 0.011  & - & 0.617  & 0.015 & - & 0.651  & 0.013 & - \\
			\bottomrule
	\end{tabular}}
\end{table*}

\subsection{Results on the TUPAC dataset}
\label{sec:result_tupac}

We then present the detection results on the TUPAC dataset. As described
in Section~\ref{sec:tupac}, the detection model was trained on the training and validation sets with the generated incomplete annotations. The candidate values of the class prior $\pi$ ranged from 0.02 to 0.07 with an increment of 0.01, and $\pi = 0.05$ was selected based on the validation set.

\begin{figure}[!t]
	\centerline{\includegraphics[width=0.95\textwidth]{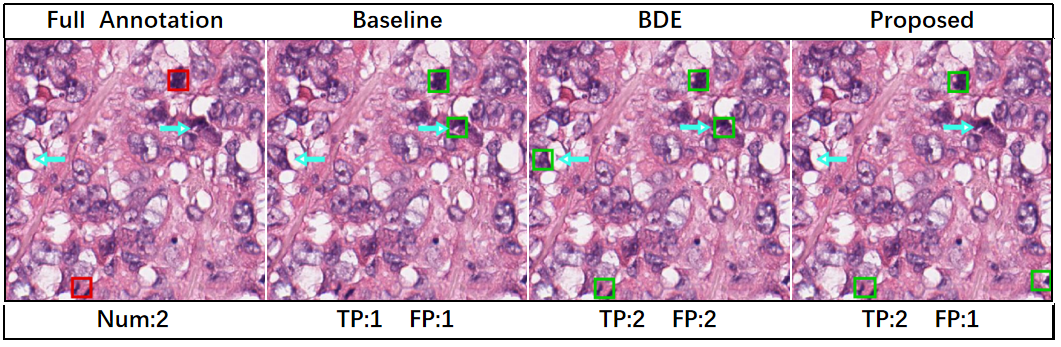}}
	\caption{An example of detection results on a test image patch for the TUPAC dataset. The gold standard full annotations and the number of annotated cells in the full annotations are also shown for the test image patch for reference. TP and FP represent the numbers of true positive and false positive detection results on the patch, respectively. Note the regions highlighted by arrows for comparison. The cyan arrows indicate examples of true negative cases given by the proposed method but labeled as positive by BDE.} 
	\label{fig:tupac}
\end{figure}

An example of the detection results of the proposed and competing methods is shown in Fig.~\ref{fig:tupac}. The bounding boxes predicted by each method are displayed, together with the complete annotations for reference. The numbers of true positive and false positive detection results on the patch are indicated for each method. Our method performs better than the competing methods with fewer false positives or more true positive boxes.

For quantitative evaluation, we computed the means and standard deviations of the recall, precision, and F1-score of the detection results on the test set for each method. Also, the upper bound performance achieved with the complete annotations for training was computed. These results are shown in Table~\ref{tab:tupac}. Compared with the competing methods, the proposed method has higher recall, precision, and F1-score, and the results of our method are closer to the upper bound. 
In addition, in Table~\ref{tab:tupac} the results of the proposed and competing methods are compared using paired Student's $t$-tests, and the improvement of our method is statistically significant. These observations indicate the better detection performance of the proposed method.

\begin{table*}[!t]
    \label{tab:tup}
	\centering
	\caption{The means and \textit{standard deviations}~(stds) of the recall, precision, and F1-score of the detection results on the test set for the TUPAC dataset. The best results are highlighted in bold. The upper bound performance is also shown for reference. Asterisks indicate that the difference between the proposed method and the competing method is significant using a paired Student's $t$-test after Benjamini-Hochberg correction for multiple comparisons. ($^{*}p<0.05$, $^{**}p<0.01$)}
	\label{tab:tupac}
	\resizebox{0.98\textwidth}{!}{
		 \begin{tabular}{c >{\centering\arraybackslash}p{1.3cm} >{\centering\arraybackslash}p{1.4cm} >{\centering\arraybackslash}p{1.5cm} >{\centering\arraybackslash}p{1.3cm} >{\centering\arraybackslash}p{1.4cm} >{\centering\arraybackslash}p{1.5cm} >{\centering\arraybackslash}p{1.3cm} >{\centering\arraybackslash}p{1.4cm} >{\centering\arraybackslash}p{1.5cm}}
			\toprule
			\multirow{2}{*}{Method} & \multicolumn{3}{c }{Recall}& \multicolumn{3}{c }{Precision}&\multicolumn{3}{c }{F1-score} \\
			\cmidrule(r){2-4} \cmidrule(r){5-7}\cmidrule(r){8-10}
			& mean &  std &  $p$  & mean  & std  & $p$  & mean  & std  & $p$      \\
			\hline
			\hline
			Baseline & 0.732 & 0.075   & **   & 0.552   & 0.063 & * &0.623&0.067&* \\ 
			BDE &0.748  & 0.079 & *  & 0.578  & 0.072 & *& 0.652  & 0.068& * \\
			 Proposed  & \textbf{0.760}  & \textbf{0.082} & -  & \textbf{0.596}  & \textbf{0.068} & -& \textbf{0.667}  & \textbf{0.064}& -\\
			 \hline
			 Upper Bound& 0.775  & 0.056 & -  & 0.654  & 0.061 & -& 0.708  & 0.057& -
			\\
			\bottomrule
	\end{tabular}}
\end{table*}

\subsection{Results on the NuCLS dataset}
\label{sec:result_nucls}

\begin{figure}[!t]
	\centerline{\includegraphics[width=0.99\textwidth]{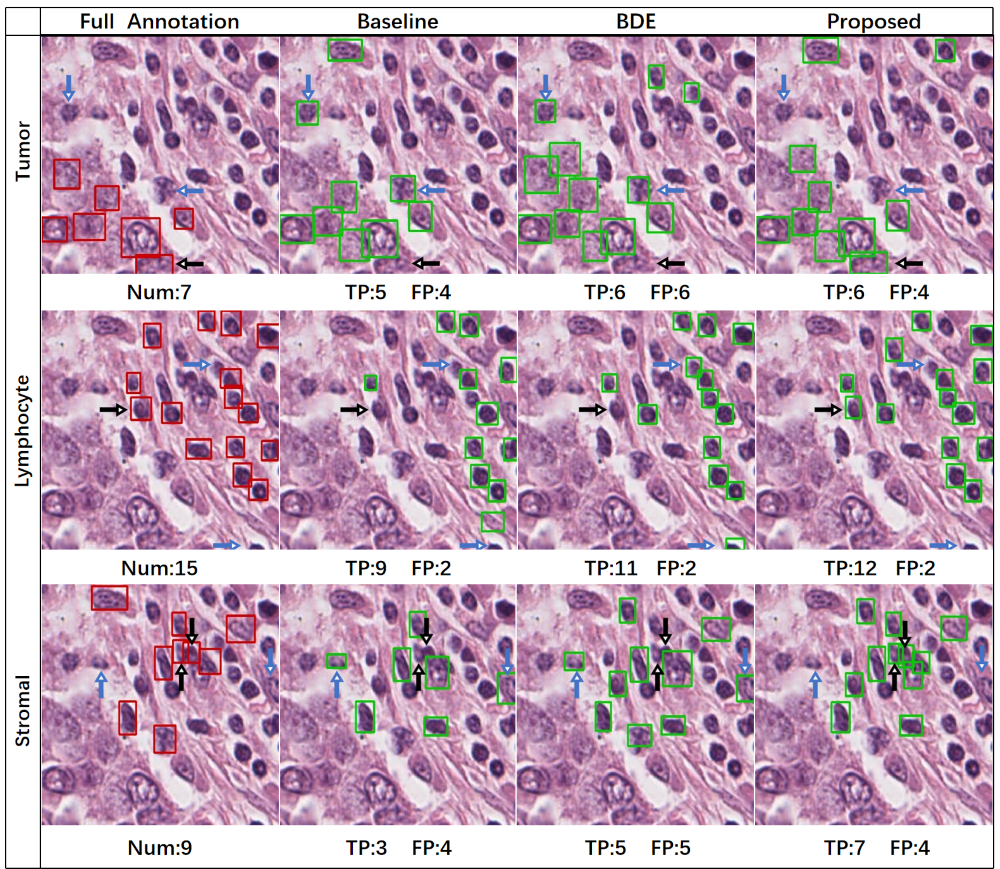}}
	\caption{Examples of representative detection results on a test image for the NuCLS dataset. The results are shown for each cell type separately. The gold standard full annotations and the numbers of annotated cells in the full annotations are also shown for the test image for reference. TP and FP represent the numbers of true positive and false positive detection results on the image, respectively. Note the regions highlighted by arrows for comparison. The black arrows indicate examples of true positive cases given by the proposed method but missed by the competing methods, whereas the blue arrows indicate examples of true negative cases given by the proposed method but labeled as positive by BDE.} 
	\label{fig:nucls}
\end{figure}

Finally, we present the results of multi-class cell detection on the NuCLS dataset.
The candidate values of the class prior $\pi_1$ (for the tumor class) ranged from 0.2 to 0.4 with an increment of 0.05, and $\pi_1=0.3$ was selected based on the validation set.

Examples of the detection results of each method on a representative test image are shown in Fig.~\ref{fig:nucls}, where the complete annotations are also shown for reference. 
Because of the large number of instances of each cell type, here the results are shown for each type separately. 
The numbers of true positive and false positive detection results on the image are indicated for each method. 
In the given examples, compared with the competing methods, for each cell type our method either produced more true positive boxes without increasing the number of false positives or produced fewer false positive boxes without decreasing the number of true positives. 

For quantitative evaluation, we computed the means and standard deviations of the recall, precision, and F1-score of the detection results
on the test set for each cell type. The results are shown in Table~\ref{tab:nucls_m_std_p}, and the upper bound performance is also given for reference. For all three cell types, the proposed method has higher recall, precision, and F1-score than the two competing approaches. 
In addition, in Table~\ref{tab:nucls_m_std_p} the proposed method is compared with the competing methods using paired Student's $t$-tests.
In most cases, the recall of the proposed method is significantly better than those of the competing methods; also, for the tumor cells the improvement of the proposed method is significant in most cases.

\begin{table*}[!t]
	
	\centering
    \arrayrulecolor{black}
	\caption{The means and \textit{standard deviations}~(stds) of the recall, precision, and F1-score of the detection results on the test set for the NuCLS dataset. The best results are highlighted in bold. The upper bound performance is also shown for reference. Asterisks indicate that the difference between the proposed method and the competing method is significant using a paired Student's $t$-test after Benjamini-Hochberg correction for multiple comparisons. ($^{*}p<0.05$, $^{**}p<0.01$, n.s. $p>0.05$)}
	\label{tab:nucls_m_std_p}
	\resizebox{0.99\textwidth}{!}{
		\begin{tabular}{
				c >{\centering\arraybackslash}p{2.8cm} >{\centering\arraybackslash}p{1.3cm} >{\centering\arraybackslash}p{1.0cm} >{\centering\arraybackslash}p{1.5cm} >{\centering\arraybackslash}p{1.3cm} >{\centering\arraybackslash}p{1.0cm} >{\centering\arraybackslash}p{1.5cm} >{\centering\arraybackslash}p{1.3cm} >{\centering\arraybackslash}p{1.0cm} >{\centering\arraybackslash}p{1.5cm}}
			\toprule
			 \multicolumn{1}{c}{\multirow{2}{*}{Cell Type}}& \multirow{2}{*}{Method}& \multicolumn{3}{c }{Recall}& \multicolumn{3}{c }{Precision}& \multicolumn{3}{c }{F1-score} \\
			\cmidrule(r){3-5} \cmidrule(r){6-8}\cmidrule(r){9-11}
			 & & mean &  std &  $p$  & mean  & std  & $p$  & mean  & std  & $p$      \\
			\hline
			\hline
			\multirow{4}{*}{Tumor}&Baseline & 0.636 & 0.041   & **   & 0.511   & 0.046 & n.s. & 0.566   & 0.044 & * \\ 
			&BDE & 0.650  & 0.058 & **  & 0.529  & 0.066 & *& 0.583  & 0.063 & ** \\
			&Proposed & \textbf{0.689} & \textbf{0.043}  & - & \textbf{0.538}  & \textbf{0.058} & -& \textbf{0.605}  & \textbf{0.053} & -\\
			\cmidrule(r){2-11}
			&Upper Bound & 0.710 & 0.041  & - & 0.622  & 0.040 & -& 0.664  & 0.041 & -\\
			\hline
			\multirow{4}{*}{Lymphocyte}&Baesline &0.526&0.087&*&0.379&0.067&n.s. &0.440&0.064&n.s. \\ 
			&BDE &0.540&0.084&n.s.&0.370&0.049&n.s.&0.438&0.055&n.s. \\
			&Proposed & \textbf{0.545} & \textbf{0.077} & - & \textbf{0.389}  & \textbf{0.046} & -& \textbf{0.454}  & \textbf{0.051} & -\\
			\cmidrule(r){2-11}
			&Upper Bound & 0.560 & 0.062  & - & 0.401  & 0.038 & - & 0.467  & 0.045 & -\\
			\hline
			\multirow{4}{*}{Stromal}&Basleine &0.388&0.046&*&0.291&0.047&n.s.&0.331&0.046&n.s.\\ 
			&BDE & 0.409&0.041&**&0.280&0.025&n.s. &0.333&0.028&n.s.\\
			&Proposed & \textbf{0.436} & \textbf{0.045}  & - & \textbf{0.293}  & \textbf{0.044} & -& \textbf{0.350}  & \textbf{0.042} & -\\
			\cmidrule(r){2-11}
			&Upper Bound & 0.412 & 0.043  & - & 0.319  & 0.031 & -& 0.359  & 0.026 & -\\
			\bottomrule
	\end{tabular}}
\end{table*}

\section{Discussion}
\label{sec:discussion}
Compared with the BDE method developed in~\cite{bde} and its extended journal version \citep{BDE_journal}, our approach addresses the problem of incomplete annotations for cell detection in histopathology images with a principled PU learning framework, and this PU learning framework has led to improved detection performance.
Note that our method and the BDE method could be complementary. Based on the density of bounding boxes, it is possible to identify additional negative samples from the unlabeled samples, which may further benefit the training of the detector, and future work could explore the integration of PU learning with the BDE method.

Because in detection problems positively labeled samples and unlabeled samples originate from the same images, in the proposed method the classification loss is approximated differently from the approximation in PU learning for classification problems. 
The results reported in Section~\ref{sec:exp_approx} confirm the benefit of the approximation we have designed for detection problems and support our discussion in Section~\ref{sec:binary}.



The proposed PU learning strategy was integrated with Faster R-CNN~\citep{Ren2015Faster} for demonstration, because it is a popular CNN-based object detector for cell detection problems~\citep{SRINIDHI2021}. In addition, we have shown that the proposed method can be applied to different backbones of Faster R-CNN, including the VGG16, ResNet50, and ResNet101 backbones.
Since the proposed method is agnostic to the architecture of the detection network, it may also be integrated with more advanced detection networks~\citep{Cai2021Cascade,Zhu2021DETR} that are recently developed, and it would be interesting to investigate in future work whether such integration can lead to improved performance.


In addition to PU learning for binary cell detection, we have extended the proposed framework to multi-class cell detection. Multi-class PU learning has also been investigated before for classification~\citep{Xu2017Multi,Shu2020}, but not for cell detection. The experimental results on the NuCLS dataset show that our method allows improved multi-class cell detection given incomplete annotations.

The problem of incomplete annotations considered in the proposed work is related to but different from semi-supervised learning. For both semi-supervised learning and the proposed work, not all cells of interest are annotated on the training images. However, they are different in terms of how the training data is obtained. 
In semi-supervised learning every cell of interest should be annotated for the labeled data. For detection problems, this can be more challenging than incomplete annotations, because it requires that experts carefully examine the annotation results to ensure that no cells of interest are left unannotated in the labeled data, whereas for incomplete annotations no such burden is required.

\section{Conclusion}
\label{sec:conclusion}

We have proposed to apply PU learning to address the problem of network training with incomplete annotations for cell detection in histopathology images. In our method, the classification loss is more appropriately computed from the incompletely annotated data during network training for both binary and multi-class cell detection. 
The experimental results on four publicly available datasets show that our method can improve the performance of cell detection in histopathology images given incomplete annotations.


\acks{This work is supported by National Natural Science Foundation of China (62001009).}

%
\ethics{The work follows appropriate ethical standards in conducting research and writing the manuscript, following all applicable laws and regulations regarding treatment of animals or human subjects.}

\coi{We declare we don't have conflicts of interest.}

\bibliography{sample}

\begin{thebibliography}{37}
\providecommand{\natexlab}[1]{#1}
\providecommand{\url}[1]{\texttt{#1}}
\expandafter\ifx\csname urlstyle\endcsname\relax
  \providecommand{\doi}[1]{doi: #1}\else
  \providecommand{\doi}{doi: \begingroup \urlstyle{rm}\Url}\fi

\bibitem[Amgad et~al.(2021)Amgad, Atteya, Hussein, Mohammed, Hafiz, Elsebaie,
  Alhusseiny, AlMoslemany, Elmatboly, Pappalardo, Sakr, Mobadersany, Rachid,
  Saad, Alkashash, Ruhban, Alrefai, Elgazar, Abdulkarim, Farag, Etman, Elsaeed,
  Alagha, Amer, Raslan, Nadim, Elsebaie, Ayad, Hanna, Gadallah, Elkady,
  Drumheller, Jaye, Manthey, Gutman, Elfandy, and Cooper]{nuclc}
Mohamed Amgad, Lamees~A. Atteya, Hagar Hussein, Kareem~Hosny Mohammed, Ehab
  Hafiz, Maha A.~T. Elsebaie, Ahmed~M. Alhusseiny, Mohamed~Atef AlMoslemany,
  Abdelmagid~M. Elmatboly, Philip~A. Pappalardo, Rokia~Adel Sakr, Pooya
  Mobadersany, Ahmad Rachid, Anas~M. Saad, Ahmad~M. Alkashash, Inas~A. Ruhban,
  Anas Alrefai, Nada~M. Elgazar, Ali Abdulkarim, Abo{-}Alela Farag, Amira
  Etman, Ahmed~G. Elsaeed, Yahya Alagha, Yomna~A. Amer, Ahmed~M. Raslan,
  Menatalla~K. Nadim, Mai A.~T. Elsebaie, Ahmed Ayad, Liza~E. Hanna, Ahmed~M.
  Gadallah, Mohamed Elkady, Bradley Drumheller, David Jaye, David Manthey,
  David~A. Gutman, Habiba Elfandy, and Lee A.~D. Cooper.
\newblock {NuCLS}: {A} scalable crowdsourcing, deep learning approach and
  dataset for nucleus classification, localization and segmentation.
\newblock \emph{arXiv preprint arXiv:2102.09099}, 2021.

\bibitem[Bertram et~al.(2020)Bertram, Veta, Marzahl, Stathonikos, Maier,
  Klopfleisch, and Aubreville]{TUPAC_alt}
Christof~A. Bertram, Mitko Veta, Christian Marzahl, Nikolas Stathonikos,
  Andreas Maier, Robert Klopfleisch, and Marc Aubreville.
\newblock Are pathologist-defined labels reproducible? comparison of the
  {TUPAC16} mitotic figure dataset with an alternative set of labels.
\newblock In \emph{Interpretable and Annotation-Efficient Learning for Medical
  Image Computing}, pages 204--213, 2020.

\bibitem[Cai et~al.(2019)Cai, Sun, Zhou, Han, and Yao]{Cai2019}
De~Cai, Xianhe Sun, Niyun Zhou, Xiao Han, and Jianhua Yao.
\newblock Efficient mitosis detection in breast cancer histology images by
  {RCNN}.
\newblock In \emph{International Symposium on Biomedical Imaging}, pages
  919--922, 2019.

\bibitem[Cai and Vasconcelos(2021)]{Cai2021Cascade}
Zhaowei Cai and Nuno Vasconcelos.
\newblock Cascade {R-CNN}: High quality object detection and instance
  segmentation.
\newblock \emph{IEEE Transactions on Pattern Analysis and Machine
  Intelligence}, 43\penalty0 (5):\penalty0 1483--1498, 2021.

\bibitem[Deng et~al.(2009)Deng, Dong, Socher, Li, Li, and Fei-Fei]{imagenet}
Jia Deng, Wei Dong, Richard Socher, Li-Jia Li, Kai Li, and Li~Fei-Fei.
\newblock Image{Net}: A large-scale hierarchical image database.
\newblock In \emph{IEEE Conference on Computer Vision and Pattern Recognition},
  pages 248--255, 2009.

\bibitem[Elkan and Noto(2008)]{Charles2008}
Charles Elkan and Keith Noto.
\newblock Learning classifiers from only positive and unlabeled data.
\newblock In \emph{International Conference on Knowledge Discovery and Data
  Mining}, pages 213--220, 2008.

\bibitem[Fusi et~al.(2013)Fusi, Metcalf, Krebs, Dive, and Blackhall]{Fusi2013}
Alberto Fusi, Robert Metcalf, Matthew Krebs, Caroline Dive, and Fiona
  Blackhall.
\newblock Clinical utility of circulating tumour cell detection in
  non-small-cell lung cancer.
\newblock \emph{Current Treatment Options in Oncology}, 14\penalty0
  (4):\penalty0 610--622, 2013.

\bibitem[Gurcan et~al.(2009)Gurcan, Boucheron, Can, Madabhushi, Rajpoot, and
  Yener]{gurcan2009histopathological}
Metin~N Gurcan, Laura~E Boucheron, Ali Can, Anant Madabhushi, Nasir~M Rajpoot,
  and Bulent Yener.
\newblock Histopathological image analysis: A review.
\newblock \emph{IEEE Reviews in Biomedical Engineering}, 2:\penalty0 147--171,
  2009.

\bibitem[He et~al.(2016)He, Zhang, Ren, and Sun]{Resnet}
Kaiming He, Xiangyu Zhang, Shaoqing Ren, and Jian Sun.
\newblock Deep residual learning for image recognition.
\newblock In \emph{IEEE Conference on Computer Vision and Pattern Recognition},
  pages 770--778, 2016.

\bibitem[He et~al.(2021)He, Zhao, and Wong]{HE2021}
Yunjie He, Hong Zhao, and Stephen T.~C. Wong.
\newblock Deep learning powers cancer diagnosis in digital pathology.
\newblock \emph{Computerized Medical Imaging and Graphics}, 88:\penalty0
  101820, 2021.

\bibitem[Kingma and Ba(2014)]{adam}
Diederik~P Kingma and Jimmy Ba.
\newblock Adam: A method for stochastic optimization.
\newblock \emph{arXiv preprint arXiv:1412.6980}, 2014.

\bibitem[Kiryo et~al.(2017)Kiryo, Niu, du~Plessis, and Sugiyama]{Kiryo2017}
Ryuichi Kiryo, Gang Niu, Marthinus~C. du~Plessis, and Masashi Sugiyama.
\newblock Positive-unlabeled learning with non-negative risk estimator.
\newblock In \emph{Advances in Neural Information Processing Systems}, pages
  1674--1684, 2017.

\bibitem[Li et~al.(2020)Li, Han, Kang, Shi, Yan, Tong, Bu, Cui, Feng, and
  Yang]{bde}
Hansheng Li, Xin Han, Yuxin Kang, Xiaoshuang Shi, Mengdi Yan, Zixu Tong, Qirong
  Bu, Lei Cui, Jun Feng, and Lin Yang.
\newblock A novel loss calibration strategy for object detection networks
  training on sparsely annotated pathological datasets.
\newblock In \emph{International Conference on Medical Image Computing and
  Computer-Assisted Intervention}, pages 320--329, 2020.

\bibitem[Li et~al.(2021)Li, Kang, Yang, Wu, Shi, Feihong, Liu, Hu, Ma, Cui,
  Feng, and Yang]{BDE_journal}
Hansheng Li, Yuxin Kang, Wentao Yang, Zhuoyue Wu, Xiaoshuang Shi, Liu Feihong,
  Jianye Liu, Lingyu Hu, Qian Ma, Lei Cui, Jun Feng, and Lin Yang.
\newblock A robust training method for pathological cellular detector via
  spatial loss calibration.
\newblock \emph{Frontiers in Medicine}, 8:\penalty0 767625, 2021.

\bibitem[Li et~al.(2019)Li, Yang, Huang, Da, Yang, Hu, Duan, Wang, and
  Li]{Li2019Signet}
Jiahui Li, Shuang Yang, Xiaodi Huang, Qian Da, Xiaoqun Yang, Zhiqiang Hu,
  Qi~Duan, Chaofu Wang, and Hongsheng Li.
\newblock Signet ring cell detection with a semi-supervised learning framework.
\newblock In \emph{International Conference on Information Processing in
  Medical Imaging}, pages 842--854, 2019.

\bibitem[Lin et~al.(2017)Lin, Goyal, Girshick, He, and
  Doll{\'a}r]{Lin2017RetinaNet}
Tsung-Yi Lin, Priya Goyal, Ross Girshick, Kaiming He, and Piotr Doll{\'a}r.
\newblock Focal loss for dense object detection.
\newblock In \emph{International Conference on Computer Vision}, pages
  2980--2988, 2017.

\bibitem[Lu et~al.(2021)Lu, Chen, Williamson, Zhao, Shady, Lipkova, and
  Mahmood]{Lu2021AI}
Ming~Y Lu, Tiffany~Y Chen, Drew F~K Williamson, Melissa Zhao, Maha Shady, Jana
  Lipkova, and Faisal Mahmood.
\newblock {AI}-based pathology predicts origins for cancers of unknown primary.
\newblock \emph{Nature}, 594\penalty0 (7861):\penalty0 106--110, 2021.

\bibitem[Marostica et~al.(2021)Marostica, Barber, Denize, Kohane, Signoretti,
  Jeffrey, and Yu]{Marostica2021}
Eliana Marostica, Rebecca Barber, Thomas Denize, Isaac~S Kohane, Sabina
  Signoretti, Golden~A Jeffrey, and Kun-Hsing Yu.
\newblock Development of a histopathology informatics pipeline for
  classification and prediction of clinical outcomes in subtypes of renal cell
  carcinoma.
\newblock \emph{Clinical Cancer Research}, 27\penalty0 (10):\penalty0
  2868--2878, 2021.

\bibitem[Neubeck and Van~Gool(2006)]{nms}
Alexander Neubeck and Luc Van~Gool.
\newblock Efficient non-maximum suppression.
\newblock In \emph{International Conference on Pattern Recognition}, pages
  850--855, 2006.

\bibitem[Noorbakhsh et~al.(2020)Noorbakhsh, Farahmand, Foroughi~Pour, Namburi,
  Caruana, Rimm, Soltanieh-ha, Zarringhalam, and Chuang]{Noorbakhsh2020}
Javad Noorbakhsh, Saman Farahmand, Ali Foroughi~Pour, Sandeep Namburi, Dennis
  Caruana, David Rimm, Mohammad Soltanieh-ha, Kourosh Zarringhalam, and
  Jeffrey~H Chuang.
\newblock Deep learning-based cross-classifications reveal conserved spatial
  behaviors within tumor histological images.
\newblock \emph{Nature Communications}, 11:\penalty0 6367, 2020.

\bibitem[Ren et~al.(2017)Ren, He, Girshick, and Sun]{Ren2015Faster}
Shaoqing Ren, Kaiming He, Ross Girshick, and Jian Sun.
\newblock Faster {R-CNN}: Towards real-time object detection with region
  proposal networks.
\newblock \emph{IEEE Transactions on Pattern Analysis and Machine
  Intelligence}, 39\penalty0 (6):\penalty0 1137--1149, 2017.

\bibitem[Roux et~al.(2013)Roux, Racoceanu, Lom{\'e}nie, Maria, Irshad, Klossa,
  Capron, Genestie, Naour, and Gurcan]{Roux2013mitos}
Ludovic Roux, Daniel Racoceanu, Nicolas Lom{\'e}nie, Kulikova Maria, Humayun
  Irshad, Jacques Klossa, Fr{\'e}d{\'e}rique Capron, Catherine Genestie,
  Le~Gilles Naour, and Metin~N Gurcan.
\newblock Mitosis detection in breast cancer histological images {An ICPR 2012
  contest}.
\newblock \emph{Journal of Pathology Informatics}, 4:\penalty0 8, 2013.

\bibitem[Shu et~al.(2020)Shu, Lin, Yan, and Li]{Shu2020}
Senlin Shu, Zhuoyi Lin, Yan Yan, and Li~Li.
\newblock Learning from multi-class positive and unlabeled data.
\newblock In \emph{International Conference on Data Mining}, pages 1256--1261,
  2020.

\bibitem[Simonyan and Zisserman(2015)]{vgg}
Karen Simonyan and Andrew Zisserman.
\newblock Very deep convolutional networks for large-scale image recognition.
\newblock \emph{arXiv preprint arXiv:1409.1556}, 2015.

\bibitem[Sirinukunwattana et~al.(2016)Sirinukunwattana, Raza, Tsang, Snead,
  Cree, and Rajpoot]{K2016Locality}
Korsuk Sirinukunwattana, Shan E~Ahmed Raza, Yee-Wah Tsang, David R.~J. Snead,
  Ian~A. Cree, and Nasir~M. Rajpoot.
\newblock Locality sensitive deep learning for detection and classification of
  nuclei in routine colon cancer histology images.
\newblock \emph{IEEE Transactions on Medical Imaging}, 35\penalty0
  (5):\penalty0 1196--1206, 2016.

\bibitem[Srinidhi et~al.(2021)Srinidhi, Ciga, and Martel]{SRINIDHI2021}
Chetan~L. Srinidhi, Ozan Ciga, and Anne~L. Martel.
\newblock Deep neural network models for computational histopathology: A
  survey.
\newblock \emph{Medical Image Analysis}, 67:\penalty0 101813, 2021.

\bibitem[Sun et~al.(2020)Sun, Huang, Molina, Dong, and Zhang]{Sun2020}
Yibao Sun, Xingru Huang, Edgar Giussepi~Lopez Molina, Le~Dong, and Qianni
  Zhang.
\newblock Signet ring cells detection in histology images with similarity
  learning.
\newblock In \emph{International Symposium on Biomedical Imaging}, pages
  490--494, 2020.

\bibitem[Tomczak et~al.(2015)Tomczak, Czerwinska, and Wiznerowicz]{TCGA}
Katarzyna Tomczak, Patrycja Czerwinska, and Maciej Wiznerowicz.
\newblock The cancer genome atlas ({TCGA}): An immeasurable source of
  knowledge.
\newblock \emph{Contemporary Oncology}, 19\penalty0 (1A):\penalty0 A68--A77,
  2015.

\bibitem[van~der Laak et~al.(2021)van~der Laak, Litjens, and Ciompi]{Laak2021}
Jeroen van~der Laak, Geert Litjens, and Francesco Ciompi.
\newblock Deep learning in histopathology: the path to the clinic.
\newblock \emph{Nature Medicine}, 27:\penalty0 775--784, 2021.

\bibitem[Veta et~al.(2014)Veta, Pluim, van Diest, and
  Viergever]{veta2014breast}
Mitko Veta, Josien P.~W. Pluim, Paul~J. van Diest, and Max~A. Viergever.
\newblock Breast cancer histopathology image analysis: A review.
\newblock \emph{IEEE Transactions on Biomedical Engineering}, 61\penalty0
  (5):\penalty0 1400--1411, 2014.

\bibitem[Veta et~al.(2019)Veta, Heng~Yujing, Stathonikos, Ehteshami, Beca,
  Wollmann, Rohr, Manan, Wang, Rousson, Hedlund, Tellez, Ciompi, Zerhouni,
  Lanyi, Viana, Kovalev, Liauchuk, Phoulady, Qaiser, Graham, Rajpoot, Sjöblom,
  Molin, Paeng, Hwang, Park, Jia, Chang, Xu, Beck, {van Diest}, and
  Pluim]{TUPAC_ori}
Mitko Veta, J.~Heng~Yujing, Nikolas Stathonikos, Babak~Bejnordi Ehteshami,
  Francisco Beca, Thomas Wollmann, Karl Rohr, A.~Shah Manan, Dayong Wang,
  Mikael Rousson, Martin Hedlund, David Tellez, Francesco Ciompi, Erwan
  Zerhouni, David Lanyi, Matheus Viana, Vassili Kovalev, Vitali Liauchuk,
  Hady~Ahmady Phoulady, Talha Qaiser, Simon Graham, Nasir Rajpoot, Erik
  Sjöblom, Jesper Molin, Kyunghyun Paeng, Sangheum Hwang, Sunggyun Park,
  Zhipeng Jia, Eric I-Chao Chang, Yan Xu, Andrew~H. Beck, Paul~J. {van Diest},
  and Josien~P.W. Pluim.
\newblock Predicting breast tumor proliferation from whole-slide images: The
  {TUPAC}16 challenge.
\newblock \emph{Medical Image Analysis}, 54:\penalty0 111--121, 2019.

\bibitem[Wang et~al.(2022)Wang, Huang, Lee, Shen, Meng, and Gaol]{Wang2022}
Ching-Wei Wang, Sheng-Chuan Huang, Yu-Ching Lee, Yu-Jie Shen, Shwu-Ing Meng,
  and Jeff~L. Gaol.
\newblock Deep learning for bone marrow cell detection and classification on
  whole-slide images.
\newblock \emph{Medical Image Analysis}, 75:\penalty0 102270, 2022.

\bibitem[Xu et~al.(2016)Xu, Xiang, Liu, Gilmore, Wu, Tang, and
  Madabhushi]{xu2015Stacked}
Jun Xu, Lei Xiang, Qingshan Liu, Hannah Gilmore, Jianzhong Wu, Jinghai Tang,
  and Anant Madabhushi.
\newblock Stacked sparse autoencoder ({SSAE}) for nuclei detection on breast
  cancer histopathology images.
\newblock \emph{IEEE Transactions on Medical Imaging}, 35\penalty0
  (1):\penalty0 119--130, 2016.

\bibitem[Xu et~al.(2017)Xu, Xu, Xu, and Tao]{Xu2017Multi}
Yixing Xu, Chang Xu, Chao Xu, and Dacheng Tao.
\newblock Multi-positive and unlabeled learning.
\newblock In \emph{International Joint Conference on Artificial Intelligence},
  pages 3182--3188, 2017.

\bibitem[Yang et~al.(2020)Yang, Liang, and Carin]{Yang2020object}
Yuewei Yang, Kevin~J Liang, and Lawrence Carin.
\newblock Object detection as a positive-unlabeled problem.
\newblock In \emph{British Machine Vision Conference}, 2020.

\bibitem[Zhao et~al.(2021)Zhao, Pang, Liu, and Ye]{miccai}
Zipei Zhao, Fengqian Pang, Zhiwen Liu, and Chuyang Ye.
\newblock Positive-unlabeled learning for cell detection in histopathology
  images with incomplete annotations.
\newblock In \emph{International Conference on Medical Image Computing and
  Computer-Assisted Intervention}, pages 509--518, 2021.

\bibitem[Zhu et~al.(2021)Zhu, Su, Lu, Li, Wang, and Dai]{Zhu2021DETR}
Xizhou Zhu, Weijie Su, Lewei Lu, Bin Li, Xiaogang Wang, and Jifeng Dai.
\newblock Deformable {DETR}: Deformable transformers for end-to-end object
  detection.
\newblock In \emph{International Conference on Learning Representations}, 2021.

\end{thebibliography}


\begin{appendices}
\appendix

\renewcommand{\appendixname}{Appendix~\Alph{section}}
\section{Baseline performance achieved with the weighted cross entropy loss or focal loss}
\label{app:loss}

\newcounter{tableA}
\setcounter{tableA}{0}
\renewcommand{\thetable}{A\arabic{tableA}}

In this appendix, we present the results of the baseline method achieved with the weighted cross entropy loss and the focal loss~\citep{Lin2017RetinaNet} for the experiment on the MITOS-ATYPIA-14 dataset with the experimental settings specified for Section~\ref{sec:result_mitos_performance}.
The weighted cross entropy loss $H_{\mathrm{wCE}}(c,z)$ and the focal loss $H_{\mathrm{focal}}(c,z)$ are defined as
\begin{eqnarray}
	\label{eq:wce}
	H_{\mathrm{wCE}}(c,z) &=& -wz\log(c)-(1-z)\log(1-c),\\
	H_{\mathrm{focal}}(c,z) &=& -\alpha z(1-c)^{\gamma}\log(c)-(1-\alpha)(1-z)c^{\gamma}\log(1-c).
	\label{eq:focal}
\end{eqnarray}
Here, $w$ is a weight for the positive class in the weighted cross entropy loss $H_{\mathrm{wCE}}(c,z)$ that alleviates the problem of class imbalance, and we set $w=\pi^{-1}$; $\alpha$ and $\gamma$ are the parameters for the focal loss $H_{\mathrm{focal}}(c,z)$ that balance the importance of positive/negative samples and easy/hard samples, respectively, and we set $\alpha = 0.25$ and $\gamma = 2$ according to \cite{Lin2017RetinaNet}.



The detection performance achieved with the weighted cross entropy loss and focal loss is summarized in Table~\ref{tab:loss}, where the means and standard deviations of the average recall, average precision, and average F1-score of the five folds are listed. Compared with the baseline results achieved with the standard cross entropy loss in Table~\ref{tab:mitos_m_std_p}, the use of the two alternative losses did not lead to improved detection performance (the F1-score is reduced).

\begin{table*}[h]
    \addtocounter{tableA}{1}
    
	\centering
	\caption{The means and \textit{standard deviations}~(stds) (mean$\pm$std) of the average recall, average precision, and average F1-score of the five folds for the MITOS-ATYPIA-14 dataset with the experimental settings specified for Section~\ref{sec:result_mitos_performance} when different losses were used for the baseline method.}
	\label{tab:loss}

	\resizebox{0.85\textwidth}{!}{
		\begin{tabular}{c >{\centering\arraybackslash}p{3.0cm} >{\centering\arraybackslash}p{3.0cm} >{\centering\arraybackslash}p{3.0cm}}
			\toprule
			 Loss & Recall &  Precision & F1-score          \\
			\hline
			\hline
			Weighted Cross Entropy     					&    0.603$\pm$0.079  		&  0.356$\pm$0.026         	&      0.445$\pm$0.034  			     \cr
			Focal 				&    0.562$\pm$0.091	&  0.394$\pm$0.032		&      0.461$\pm$0.048  	\\
						\bottomrule
	\end{tabular}}
	
\end{table*}

\section{Sensitivity to the class prior $\pi$}
\label{app:pi}

\newcounter{tableB}
\setcounter{tableB}{0}
\renewcommand{\thetable}{B\arabic{tableB}}

The evaluation of the sensitivity of the detection performance to the class prior $\pi$ is provided in this appendix.
We computed the F1-score corresponding to each candidate $\pi$ on the test set for the MITOS-ATYPIA-14 dataset with the experimental settings specified for Section~\ref{sec:result_mitos_performance}, and the results are shown in Table~\ref{tab:mitos_f1_pi} for each fold. The difference between the best and second best results achieved with different $\pi$ values is small, and both of them are better than the BDE results in Table~\ref{tab:mitos_f1_fold}.

\begin{table*}[!t]
    \addtocounter{tableB}{1}

	\centering
	\caption{The average F1-score of the detection results  corresponding to each $\pi$ on the MITOS-ATYPIA-14 dataset (with the experimental settings specified for Section~\ref{sec:result_mitos_performance}). The result associated with the $\pi$ value selected based on the validation set is highlighted in bold.}
	\label{tab:mitos_f1_pi}
	\resizebox{0.65\textwidth}{!}{
		\begin{tabular}{c >{\centering\arraybackslash}p{1.5cm} >{\centering\arraybackslash}p{1.5cm} >{\centering\arraybackslash}p{1.5cm} >{\centering\arraybackslash}p{1.5cm} >{\centering\arraybackslash}p{1.5cm} }
			\toprule
			\multirow{2}{*}{$\pi$}& \multicolumn{5}{c }{F1-score} \\
			\cmidrule(r){2-6} 
			&    Fold 1 & Fold 2        &      Fold 3 &  Fold 4     &   Fold 5            \\
			\hline
			\hline
			$0.025$ 				&    0.508  		&  0.465          	&      0.460  			&   0.431       		&    0.549     \cr
			$0.030$      					&    0.517  		&  0.480          	&      0.471  			&   0.447       		&    0.560     \cr
			$0.035$ 				&    0.523	&  \bf{0.487}		&      0.470  	&   0.456    &    0.572 \cr
			$0.040$ 				&    \bf{0.524}	&  0.485		&      0.488  	&   \bf{0.457}    &    0.587 \cr
			$0.045$ 				&    0.514	&  0.475		&      \bf{0.484}  	&   0.443    &    \bf{0.583} \cr
			$0.050$ 				&    0.507	&  0.468		&      0.473  	&   0.445    &    0.580 \\
						\bottomrule
	\end{tabular}}
\end{table*}

\section{The impact of random effects}
\label{app:random}
\newcounter{tableC}
\setcounter{tableC}{0}
\renewcommand{\thetable}{C\arabic{tableC}}

\begin{table*}[!t]
    
    \addtocounter{tableC}{1}
	\centering
	\caption{The means and \textit{standard deviations}~(stds) (mean$\pm$std) of the recall, precision, and F1-score results of five independent runs achieved on the MITOS-ATYPIA-14 dataset (with the experimental settings specified for Section~\ref{sec:result_mitos_performance} and the first fold).}
	\label{tab:repeat}
	\resizebox{0.75\textwidth}{!}{
		\begin{tabular}{c >{\centering\arraybackslash}p{3.0cm} >{\centering\arraybackslash}p{3.0cm} >{\centering\arraybackslash}p{3.0cm}}
			\toprule
			 Method & Recall &  Precision & F1-score          \\
			\hline
			\hline
			Baseline     					&    0.605$\pm$0.006  		&  0.412$\pm$0.005         	&      0.491$\pm$0.002  			     \cr
			BDE      					&    0.639$\pm$0.003  		&  0.431$\pm$0.006          	&      0.515$\pm$0.004  			     \cr
			Proposed 				&    0.642$\pm$0.006	&  0.441$\pm$0.006		&      0.525$\pm$0.005  	\\
						\bottomrule
	\end{tabular}}
\end{table*}

During network training, random effects such as batch selection can lead to different learned network weights. 
Therefore, we investigated the impact of random effects on the detection performance for the proposed and competing methods using the MITOS-ATYPIA-14 dataset with the experimental settings specified for Section~\ref{sec:result_mitos_performance}.
The baseline method, BDE method, and proposed method were repeated independently five times with the first fold (including the results presented in Section~\ref{sec:result_mitos_performance}). The means and standard deviations of the recall, precision, and F1-score results of the five runs are shown in Table~\ref{tab:repeat}. 
The standard deviations are relatively small compared with the means, indicating that all methods are robust to random effects, and our method is better than the competing methods with higher recall, precision, and F1-score.
\end{appendices}

\end{document}